\documentclass[10pt,twocolumn,letterpaper]{article}

\usepackage[pagenumbers]{cvpr} 

\usepackage{graphicx}
\usepackage{amsmath}
\usepackage{amssymb}
\usepackage{booktabs}
\usepackage{tablefootnote}
\usepackage[flushleft]{threeparttable}

\usepackage{lipsum}
\usepackage{bbding}
\usepackage{url}
\usepackage[accsupp]{axessibility}  

\usepackage{color}
\usepackage{enumitem}
\definecolor{mypink3}{cmyk}{0, 0.7808, 0.4429, 0.1412}
\definecolor{myblue2}{cmyk}{0, 0., 0.4412, 0.4808}
\definecolor{myblue}{cmyk}{0, 0.7808, 0., 0.1412}

\usepackage[pagebackref,breaklinks,colorlinks]{hyperref}

\usepackage[capitalize]{cleveref}
\crefname{section}{Sec.}{Secs.}
\Crefname{section}{Section}{Sections}
\Crefname{table}{Table}{Tables}
\crefname{table}{Tab.}{Tabs.}

\def\cvprPaperID{2825} 
\def\confName{CVPR}
\def\confYear{2023}

\begin{document}

\title{Spherical Transformer for LiDAR-based 3D Recognition}

\author{Xin Lai$^{1}$\hspace{1.0cm}Yukang Chen$^{1}$\hspace{1.0cm}Fanbin Lu$^{1}$\hspace{1.0cm}Jianhui Liu$^{2}$\hspace{1.0cm}Jiaya Jia$^{1,3}$\\
$^{1}$The Chinese University of Hong Kong~~~
$^{2}$The University of Hong Kong~~~
$^{3}$SmartMore~~~
}

\maketitle

\begin{abstract}
LiDAR-based 3D point cloud recognition has benefited various applications. Without specially considering the LiDAR point distribution, most current methods suffer from information disconnection and limited receptive field, especially for the sparse distant points. In this work, we study the varying-sparsity distribution of LiDAR points and present \textbf{SphereFormer} to directly aggregate information from dense close points to the sparse distant ones. We design radial window self-attention that partitions the space into multiple non-overlapping narrow and long windows. It overcomes the disconnection issue and enlarges the receptive field smoothly and dramatically, which significantly boosts the performance of sparse distant points. Moreover, to fit the narrow and long windows, we propose exponential splitting to yield fine-grained position encoding and dynamic feature selection to increase model representation ability. Notably, our method ranks 1\textsuperscript{st} on both nuScenes and SemanticKITTI semantic segmentation benchmarks with $81.9\%$ and $74.8\%$ mIoU, respectively. Also, we achieve the 3\textsuperscript{rd} place on nuScenes object detection benchmark with $72.8\%$ NDS and $68.5\%$ mAP. Code is available at \url{https://github.com/dvlab-research/SphereFormer.git}.
\end{abstract}

\section{Introduction}
\label{sec:intro}

\begin{figure*}
\begin{center}
\includegraphics[width=1.0\linewidth]{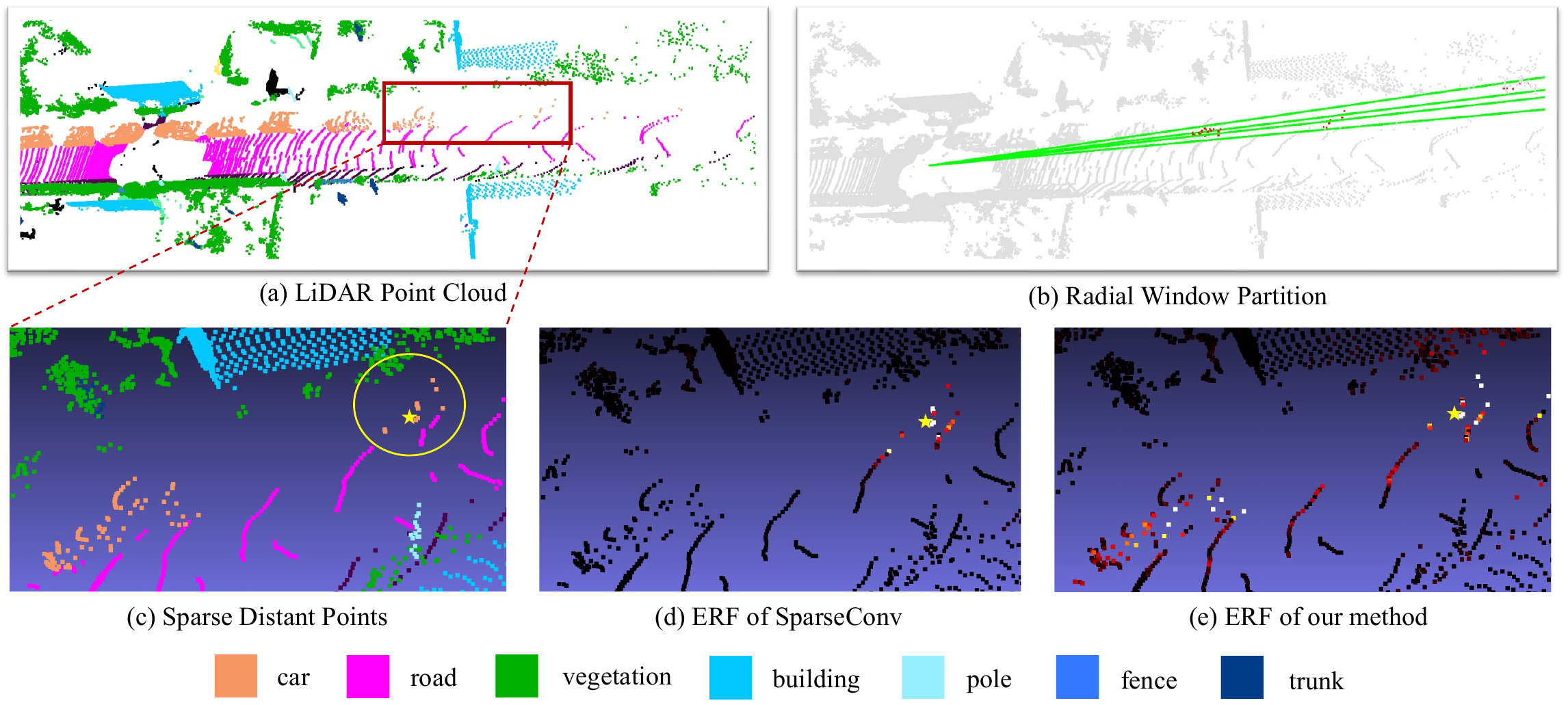}
\end{center}
\vspace{-0.6cm}
\caption{Effective Receptive Field (ERF) of SparseConv and ours. (a) LiDAR point cloud. (b) Radial window partition. Only a single radial window is shown. Points inside the window are marked in red. (c) Zoom-in sparse distant points. A sparse \textit{car} is circled in yellow. (d) ERF of SparseConv, given the point of interest (with yellow star). White and red denote high contribution. (e) ERF of ours.}
\label{fig:erf}
\end{figure*}

Nowadays, point clouds can be easily collected by LiDAR sensors. They are extensively used in various industrial applications, such as autonomous driving and robotics. In contrast to 2D images where pixels are arranged densely and regularly, LiDAR point clouds possess the varying-sparsity property --- points near the LiDAR are quite dense, while points far away from the sensor are much sparser, as shown in Fig.~\ref{fig:erf} (a).

\begin{figure}
\begin{center}
\includegraphics[width=1.0\linewidth]{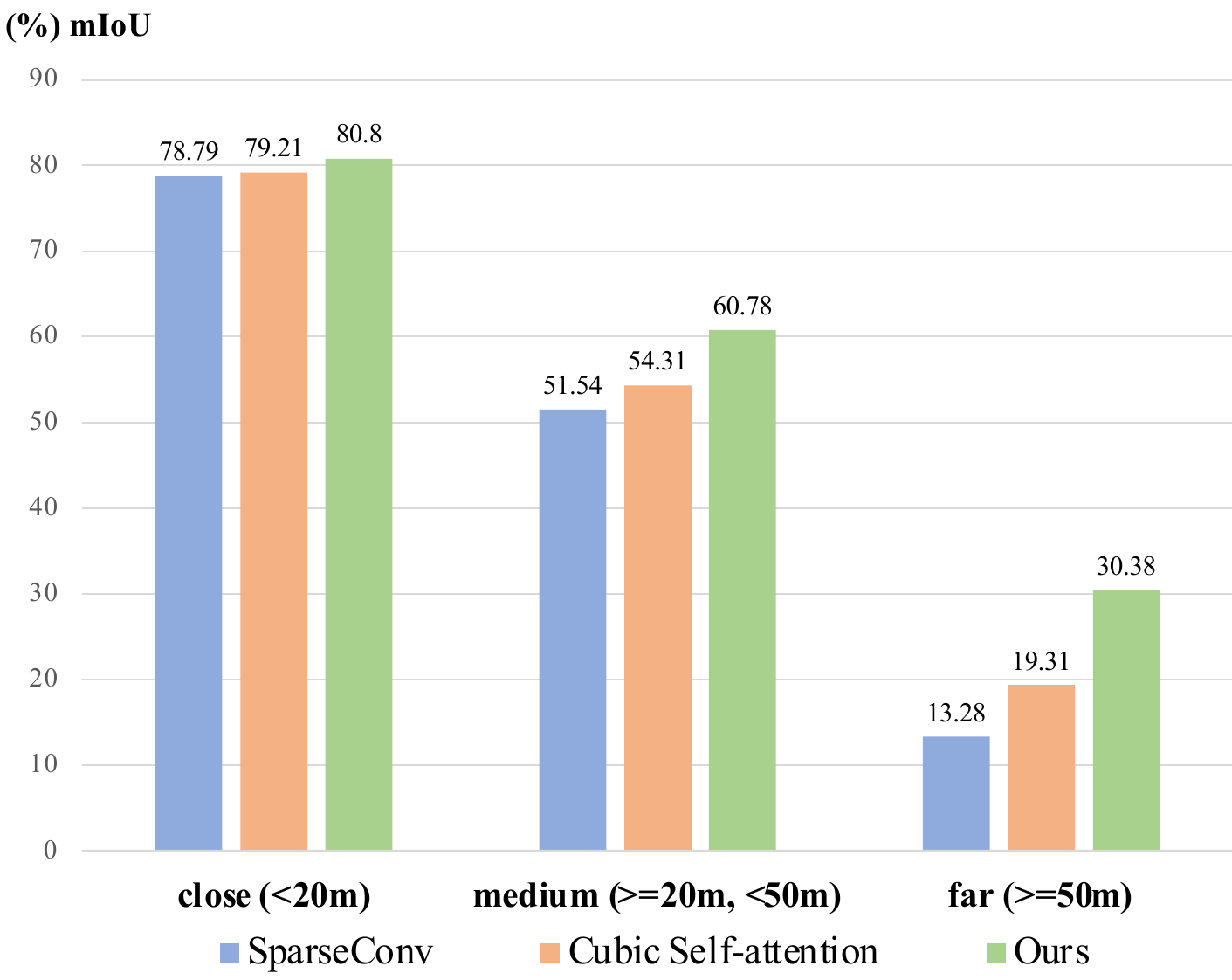}
\end{center}
\vspace{-0.4cm}
\caption{Semantic segmentation performance on nuScenes \textit{val} set for points at different distances.}
\label{fig:histogram}
\vspace{-0.3cm}
\end{figure}

However, most existing work~\cite{3DSemanticSegmentationWithSubmanifoldSparseConvNet,SubmanifoldSparseConvNet,choy20194d,yan2021sparse,tang2020searching,cheng20212,xu2021rpvnet,yan20222dpass} does not specially consider the the varying-sparsity point distribution of outdoor LiDAR point clouds. They inherit from 2D CNNs or 3D indoor scenarios, and conduct local operators (\eg, SparseConv~\cite{3DSemanticSegmentationWithSubmanifoldSparseConvNet,SubmanifoldSparseConvNet}) uniformly for all locations. This causes inferior results for the sparse distant points. As shown in Fig.~\ref{fig:histogram}, although decent performance is yielded for the dense close points, it is difficult for these methods to deal with the \textit{sparse distant points} optimally. 


We note that the root cause lies in limited receptive field. For sparse distant points, there are few surrounding neighbors. This not only results in inconclusive features, but also hinders enlarging receptive field due to information disconnection. To verify this finding, we visualize the Effective Receptive Field (ERF)~\cite{luo2016understanding} of the given feature (shown with the yellow star) in Fig.~\ref{fig:erf} (d). The ERF cannot be expanded due to disconnection, which is caused by the extreme sparsity of the distant \textit{car}.

Although window self-attention~\cite{lai2022stratified,fan2022embracing}, dilated self-attention~\cite{mao2021voxel}, and large-kernel CNN~\cite{chen2022scaling} have been proposed to conquer the limited receptive field, these methods do not specially deal with LiDAR point distribution, and remain to enlarge receptive field by stacking local operators as before, leaving the information disconnection issue still unsolved. As shown in Fig.~\ref{fig:histogram}, the method of cubic self-attention brings a limited improvement.


In this paper, we take a new direction to {\it aggregate long-range information directly in a single operator} to suit the varying-sparsity point distribution. We propose the module of \textit{SphereFormer} to perceive useful information from points 50+ meters away and yield large receptive field for feature extraction. Specifically, we represent the 3D space using spherical coordinates $(r, \theta, \phi)$ with the sensor being the origin, and partition the scene into multiple non-overlapping windows. Unlike the cubic window shape, we design radial windows that are long and narrow. They are obtained by partitioning only along the $\theta$ and $\phi$ axis, as shown in Fig.~\ref{fig:erf} (b). It is noteworthy that we make it a plugin module to conveniently insert into existing mainstream backbones. 

The proposed module does not rely on stacking local operators to expand receptive field, thus avoiding the disconnection issue, as shown in Fig.~\ref{fig:erf} (e). Also, it facilitates the sparse distant points to aggregate information from the dense-point region, which is often semantically rich. So, the performance of the distant points can be improved significantly (\ie, +17.1\% mIoU) as illustrated in Fig.~\ref{fig:histogram}. 

Moreover, to fit the long and narrow radial windows, we propose \textit{exponential splitting} to obtain fine-grained relative position encoding. The radius $r$ of a radial window can be over 50 meters, which causes large splitting intervals. It thus results in coarse position encoding when converting relative positions into integer indices. Besides, to let points at varying locations treat local and global information differently, we propose \textit{dynamic feature selection} to make further improvements.

In total, our contribution is three-fold.
\begin{itemize}
    \item We propose SphereFormer to directly aggregate long-range information from dense-point region. It increases the receptive field smoothly and helps improve the performance of \textit{sparse distant points}.
    
    \item To accommodate the radial windows, we develop exponential splitting for relative position encoding. Our dynamic feature selection further boosts performance.
    
    \item Our method achieves new state-of-the-art results on multiple benchmarks of both semantic segmentation and object detection tasks.
\end{itemize}

\section{Related Work}

\subsection{LiDAR-based 3D Recognition}
\paragraph{Semantic Segmentation.}
Segmentation~\cite{ronneberger2015u,zhao2017pyramid,chen2018encoder,lai2021semi,lai2022decouplenet,tian2022adaptive,tian2023learning,tian2022generalized,chu2022twist,icm-3d,li2021simultaneous} is a fundamental task for vision perception. Approaches for LiDAR-based semantic segmentation can be roughly grouped into three categories, \ie, view-based, point-based, and voxel-based methods. View-based methods either transform the LiDAR point cloud into a range view~\cite{wu2019squeezesegv2,xu2020squeezesegv3,behley2019semantickitti,milioto2019rangenet++,razani2021lite}, or use a bird-eye view (BEV)~\cite{zhang2020polarnet} for a 2D network to perform feature extraction. 3D geometric information is simplified. 

Point-based methods~\cite{qi2017pointnet, qi2017pointnet++, thomas2019kpconv, yan2020pointasnl, hu2020randla, tatarchenko2018tangent, lai2022stratified} adopt the point features and positions as inputs, and design abundant operators to aggregate information from neighbors. Moreover, the voxel-based solutions~\cite{3DSemanticSegmentationWithSubmanifoldSparseConvNet, SubmanifoldSparseConvNet, choy20194d} divide the 3D space into regular voxels and then apply sparse convolutions. Further, methods of \cite{yan2021sparse,tang2020searching,zhu2021cylindrical,cheng20212,liu2022less,jiang2021guided,cohen2018spherical} propose various structures for improved effectiveness. All of them focus on capturing local information. We follow this line of research, and propose to directly aggregate long-range information.

Recently, RPVNet~\cite{xu2021rpvnet} combines the three modalities by feature fusion. Furthermore, 2DPASS~\cite{yan20222dpass} incorporates 2D images during training, and \cite{Robert_2022_CVPR} fuses multi-modal features. Despite extra 2D information, the performance of these methods still lags behind compared to ours.

\paragraph{Object Detection.} 
3D object detection frameworks can be roughly categorized into single-stage~\cite{3dssd, sessd, sassd, cia-ssd, spatial-pruned-conv, chen2023voxenext} and two-stage~\cite{point-rcnn, voxel-rcnn, pvrcnn, pyramid-rcnn} methods. VoxelNet~\cite{voxelnet} extracts voxel features by PointNet~\cite{qi2017pointnet} and applies RPN~\cite{fasterrcnn} to obtain the proposals. SECOND~\cite{second} is efficient thanks to the accelerated sparse convolutions. VoTr~\cite{mao2021voxel} applies cubic window attention to voxels. LiDARMultiNet~\cite{ye2022lidarmultinet} unifies semantic segmentation, panoptic segmentation, and object detection into a single multi-task network with multiple types of supervision.  Our experiments are based on CenterPoint~\cite{yin2021center}, which is a widely used anchor-free framework. It is effective and efficient. We aim to enhance the features of sparse distant points, and our proposed module can be conveniently inserted into existing frameworks. 


\subsection{Vision Transformer}

Recently, Transformer~\cite{vaswani2017attention} become popular in various 2D image understanding tasks~\cite{dosovitskiy2020vit, pmlr-v139-touvron21a, touvron2021cait, wang2021pyramid, wang2021pvtv2, liu2021Swin, chu2021Twins, yang2021focal, dong2021cswin, vip, detr, zhu2020deformable, mao2021voxel, zhao2020san}. ViT~\cite{dosovitskiy2020vit} tokenizes every image patch and adopts a Transformer encoder to extract features. Further, PVT~\cite{wang2021pyramid} presents a hierarchical structure to obtain a feature pyramid for dense prediction. It also proposes Spatial Reduction Attention to save memory. Also, Swin Transformer~\cite{liu2021Swin} uses window-based attention and proposes the shifted window operation in the successive Transformer block. Moreover, methods of~\cite{chu2021Twins, yang2021focal, dong2021cswin} propose different designs to incorporate long-range dependencies. There are also methods~\cite{zhao2021point,mao2021voxel,lai2022stratified,fan2022embracing,sun2022swformer} that apply Transformer into 3D vision. Few of them consider the point distribution of LiDAR point cloud. In our work, we utilize the varying-sparsity property, and design radial window self-attention to capture long-range information, especially for the sparse distant points.

\section{Our Method}

In this section, we first elaborate on radial window partition in Sec.~\ref{sec:radial}. Then, we propose the improved position encoding and dynamic feature selection in Sec.~\ref{sec:pe} and \ref{sec:dfs}.





\subsection{Spherical Transformer}
\label{sec:radial}

\begin{figure}
\begin{center}
\includegraphics[width=1.0\linewidth]{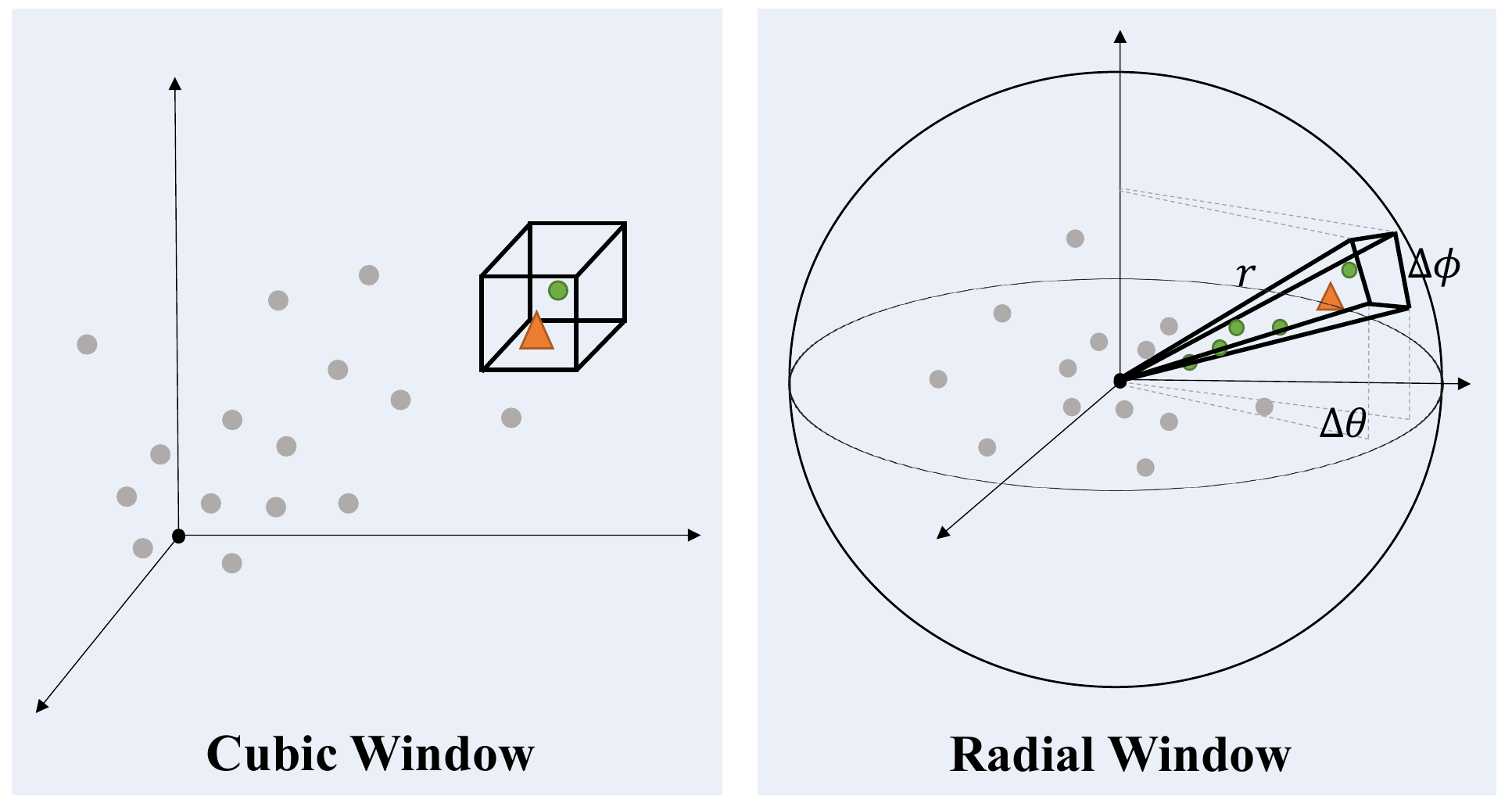}
\end{center}
\vspace{-0.6cm}
\caption{Cubic vs. Radial window partition. The radial window can directly harvest information from the dense-point region, especially for the sparse distant points.}
\label{fig:radial}
\vspace{-0.2cm}
\end{figure}

To model the long-range dependency, we adopt the window-attention~\cite{liu2021Swin} paradigm. However, unlike the cubic window attention~\cite{mao2021voxel,fan2022embracing,lai2022stratified}, we take advantage of the varying-sparsity property of LiDAR point cloud and present the SphereFormer module, as shown in Fig.~\ref{fig:radial}.

\paragraph{Radial Window Partition.} Specifically, we represent LiDAR point clouds using the spherical coordinate system $(r, \theta, \phi)$ with the LiDAR sensor being the origin. We partition the 3D space along the $\theta$ and $\phi$ axis. We, thus, obtain a number of non-overlapping radial windows with a long and narrow 'pyramid' shape, as shown in Fig.~\ref{fig:radial}. We obtain the window index for the token at ($r_i$, $\theta_i$, $\phi_i$) as
\begin{equation}
\footnotesize
\label{eq:window_partition}
    win\_index_{i} = (\lfloor \frac{\theta_i}{\Delta \theta} \rfloor, \lfloor \frac{\phi_i}{\Delta \phi} \rfloor),
\end{equation}
where $\Delta \theta$ and $\Delta \phi$ denote the window size corresponding to the $\theta$ and $\phi$ dimension, respectively.

Tokens with the same window index would be assigned to the same window. The multi-head self-attention~\cite{vaswani2017attention} is conducted within each window independently as follows.
\begin{equation}
\footnotesize
\label{eq:qkv_proj}
\mathbf{\hat{q}} = \mathbf{f} \cdot \mathbf{W}_q, \quad \mathbf{\hat{k}} = \mathbf{f} \cdot \mathbf{W}_k, \quad \mathbf{\hat{v}} = \mathbf{f} \cdot \mathbf{W}_v,
\end{equation}
where $\mathbf{f}\in \mathbb{R}^{n\times c}$ denotes the input features of a window, $\mathbf{W}_q, \mathbf{W}_k, \mathbf{W}_v \in \mathbb{R}^{c\times c}$ are the linear projection weights, and $\mathbf{\hat{q}}, \mathbf{\hat{k}}, \mathbf{\hat{v}} \in \mathbb{R}^{n\times c}$ are the projected features. Then, we split the projected features $\mathbf{\hat{q}}, \mathbf{\hat{k}}, \mathbf{\hat{v}}$ into $h$ heads (\ie, $\mathbb{R}^{n\times (h \times d)}$), and reshape them as $\mathbf{q}, \mathbf{k}, \mathbf{v} \in \mathbb{R}^{h\times n \times d}$. For each head, we perform dot product and weighted sum as
\begin{equation}
\footnotesize
\label{eq:attn_softmax}
\mathbf{attn}_k = \mathbf{softmax}(\mathbf{q}_k \cdot \mathbf{k}_k^T),
\vspace{-0.4cm}
\end{equation}
\begin{equation}
\footnotesize
\hat{\mathbf{z}}_k = \mathbf{attn}_k \cdot \mathbf{v}_{k},
\vspace{0.1cm}
\end{equation}
where $\mathbf{q}_k, \mathbf{k}_k, \mathbf{v}_k \in \mathbb{R}^{n\times d}$ denote the features of the $k$-th head, and $\mathbf{attn}_k \in \mathbb{R}^{n\times n}$ is the corresponding attention weight. Finally, we concatenate the features from all heads and apply the final linear projection with weight $\mathbf{W}_{proj} \in \mathbb{R}^{c\times c}$ to yield the output $\mathbf{z}\in \mathbb{R}^{n\times c}$ as
\begin{equation}
\footnotesize
\label{eq:concat}
\hat{\mathbf{z}} = \mathbf{concat}(\{\hat{\mathbf{z}}_0, \hat{\mathbf{z}}_1, ..., \hat{\mathbf{z}}_{h-1}\}).
\vspace{-0.3cm}
\end{equation}
\begin{equation}
\footnotesize
\mathbf{z} = \hat{\mathbf{z}} \cdot \mathbf{W}_{proj}.
\vspace{0.1cm}
\end{equation}

SphereFormer serves as a plugin module and can be conveniently inserted into existing mainstream models, \eg, SparseConvNet~\cite{3DSemanticSegmentationWithSubmanifoldSparseConvNet,SubmanifoldSparseConvNet}, MinkowskiNet~\cite{choy20194d}, local window self-attention~\cite{mao2021voxel,lai2022stratified,fan2022embracing}. In this paper, we find that inserting it into the end of each stage works well, and the network structure is given in the supplementary material. The resulting model can be applied to various downstream tasks, such as semantic segmentation and object detection, with strong performance as produced in experiments.

SphereFormer is effective for the sparse distant points to get long-range information from the dense-point region. Therefore, the sparse distant points overcome the disconnection issue, and increase the effective receptive field.

\paragraph{Comparison with Cylinder3D.}
Although both Cylinder3D~\cite{zhu2021cylindrical} and ours use polar or spherical coordinates to match LiDAR point distribution, there are two essential differences yet. First, Cylinder3D aims at a more balanced point distribution, while our target is to enlarge the receptive field smoothly and enable the sparse distant points to directly aggregate long-range information from the dense-point region. Second, what Cylinder3D does is replace the cubic voxel shape with the fan-shaped one. It remains to use local neighbors as before and still suffers from limited receptive field for the sparse distant points. Nevertheless, our method changes the way we find neighbors in a single operator (\ie, self-attention) and it is not limited to local neighbors. It thus avoids information separation between near and far objects and connects them in a natural way. 


\subsection{Position Encoding} 
\label{sec:pe}

\begin{figure}
\begin{center}
\includegraphics[width=1.0\linewidth]{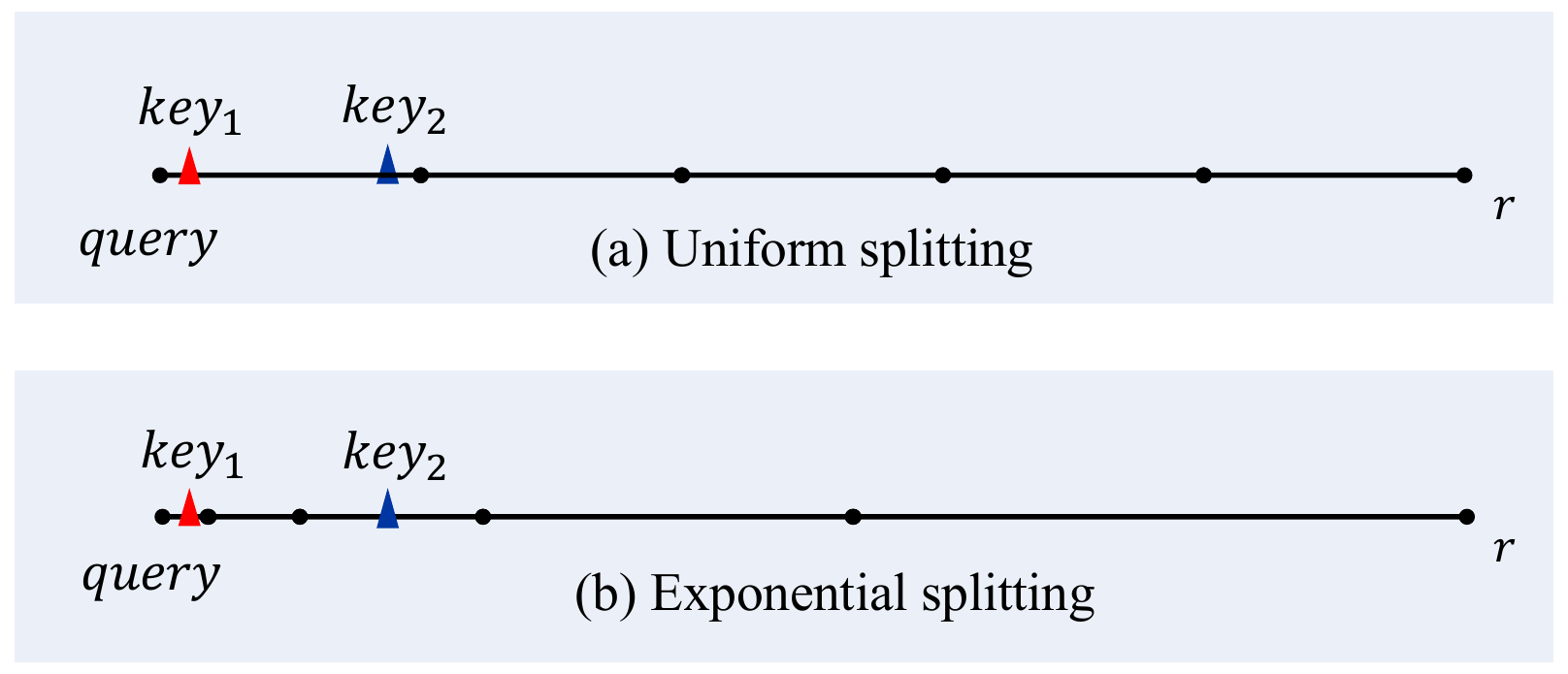}
\end{center}
\vspace{-0.6cm}
\caption{Comparison between (a) uniform splitting and (b) {exponential splitting}. The $query$ is at the leftmost point.}
\label{fig:splitting}
\vspace{-0.1cm}
\end{figure}

For the 3D point cloud network, the input features have already incorporated the absolute $xyz$ position. Therefore, there is no need to apply absolute position encoding. Also, we notice that Stratified Transformer~\cite{lai2022stratified} develops the contextual relative position encoding. It splits a relative position into several discrete parts uniformly, which converts the continuous relative positions into integers to index the positional embedding tables. 

This method works well with local cubic windows. But in our case, the radial window is narrow and long, and its radius $r$ can take even more than 50 meters, which could cause large intervals during discretization and thus coarse-grained positional encoding. As shown in Fig.~\ref{fig:splitting} (a), because of the large interval, $key_1$ and $key_2$ correspond to the same index. But there is still a considerable distance between them.

\paragraph{Exponential Splitting.} Specifically, since the $r$ dimension covers long distances, we propose \textit{exponential splitting} for the $r$ dimension as shown in Fig.~\ref{fig:splitting} (b). The splitting interval grows exponentially when the index increases. In this way, the intervals near the $query$ are much smaller, and the $key_1$ and $key_2$ can be assigned to different position encodings. Meanwhile, we remain to adopt the \textit{uniform splitting} for the $\theta$ and $\phi$ dimensions. In notation, we have a query token $q_i$ and a key token $k_j$. Their relative position $(r_{ij}, \theta_{ij}, \phi_{ij})$ is converted into integer index $(\mathbf{idx}^r_{ij}, \mathbf{idx}^\theta_{ij}, \mathbf{idx}^\phi_{ij})$ as
\begin{equation}
\begin{footnotesize}
\begin{aligned}
\mathbf{idx}^r_{ij} = \left \{
\begin{array}{lcl}
    -\max(0, \lceil \log_2(\frac{-r_{ij}}{a}) \rceil) - 1 &  & r_{ij} < 0 \\
    0 & & r_{ij} = 0 \\
    \max(0, \lceil \log_2(\frac{r_{ij}}{a}) \rceil) &  & r_{ij} > 0 \\
\end{array} \right.,\nonumber
\end{aligned}
\end{footnotesize}
\end{equation}
\begin{equation}
\begin{footnotesize}
\begin{aligned}
\mathbf{idx}^\theta_{ij} = \lfloor \frac{\theta_{ij}}{\mathbf{inteval}_\theta} \rfloor, \quad \mathbf{idx}^\phi_{ij} = \lfloor \frac{\phi_{ij}}{\mathbf{inteval}_\phi} \rfloor,\nonumber
\end{aligned}
\end{footnotesize}
\end{equation}
\begin{equation}
\begin{footnotesize}
\begin{aligned}
    \mathbf{idx}^{x} = \mathbf{idx}^{x} + \frac{L}{2},\quad x \in \{r, \theta, \phi\},\nonumber
\end{aligned}
\end{footnotesize}
\end{equation}
where $a$ is a hyper-parameter to control the starting splitting interval, and $L$ is the length of the positional embedding tables. Note that we also add the indices with $\frac{L}{2}$ to make sure they are non-negative.

The above indices ($\mathbf{idx}^r_{ij}, \mathbf{idx}^\theta_{ij}, \mathbf{idx}^\phi_{ij}$) are then used to index their positional embedding tables $\mathbf{t}_r, \mathbf{t}_\theta, \mathbf{t}_\phi \in \mathbb{R}^{L\times (h \times d)}$ to find the corresponding position encoding $\mathbf{p}^r_{ij}, \mathbf{p}^\theta_{ij}, \mathbf{p}^\phi_{ij} \in \mathbb{R}^{h\times d}$, respectively. Then, we sum them up to yield the resultant positional encoding $\mathbf{p} \in \mathbb{R}^{h\times d}$, which then performs dot product with the features of $q_{i}$ and $k_{j}$, respectively. The original Eq.~(\ref{eq:attn_softmax}) is updated to
\begin{equation}
\begin{footnotesize}
\begin{aligned}
 \mathbf{p} & = \mathbf{p}^r_{ij} + \mathbf{p}^\theta_{ij} + \mathbf{p}^\phi_{ij}, \nonumber \\
 \mathbf{pos\_bias}_{k, i, j} & = \mathbf{q}_{k,i} \cdot \mathbf{p}_{k}^T + \mathbf{k}_{k,j} \cdot \mathbf{p}_{k}^T, \\
 \mathbf{attn}_k & = \mathbf{softmax}(\mathbf{q}_k \cdot \mathbf{k}_k^T + \mathbf{pos\_bias}_k),
\end{aligned}
\end{footnotesize}
\end{equation}
where $\mathbf{pos\_bias} \in \mathbb{R}^{h\times n \times n}$ is the positional bias to the attention weight, $\mathbf{q}_{k,i} \in \mathbb{R}^{d}$ means the the $k$-th head of the $i$-th query feature, and $\mathbf{p}_k \in \mathbb{R}^{d}$ is the $k$-th head of the position encoding $\mathbf{p}$.

The \textit{exponential splitting} strategy provides smaller splitting intervals for near token pairs and larger intervals for distant ones. This operation enables a fine-grained position representation between near token pairs, and still maintains the same number of intervals in the meanwhile. Even though the splitting intervals become larger for distant token pairs, this solution actually works well since distant token pairs require less fine-grained relative position.

\subsection{Dynamic Feature Selection}
\label{sec:dfs}

\begin{figure}
\begin{center}
\includegraphics[width=1.0\linewidth]{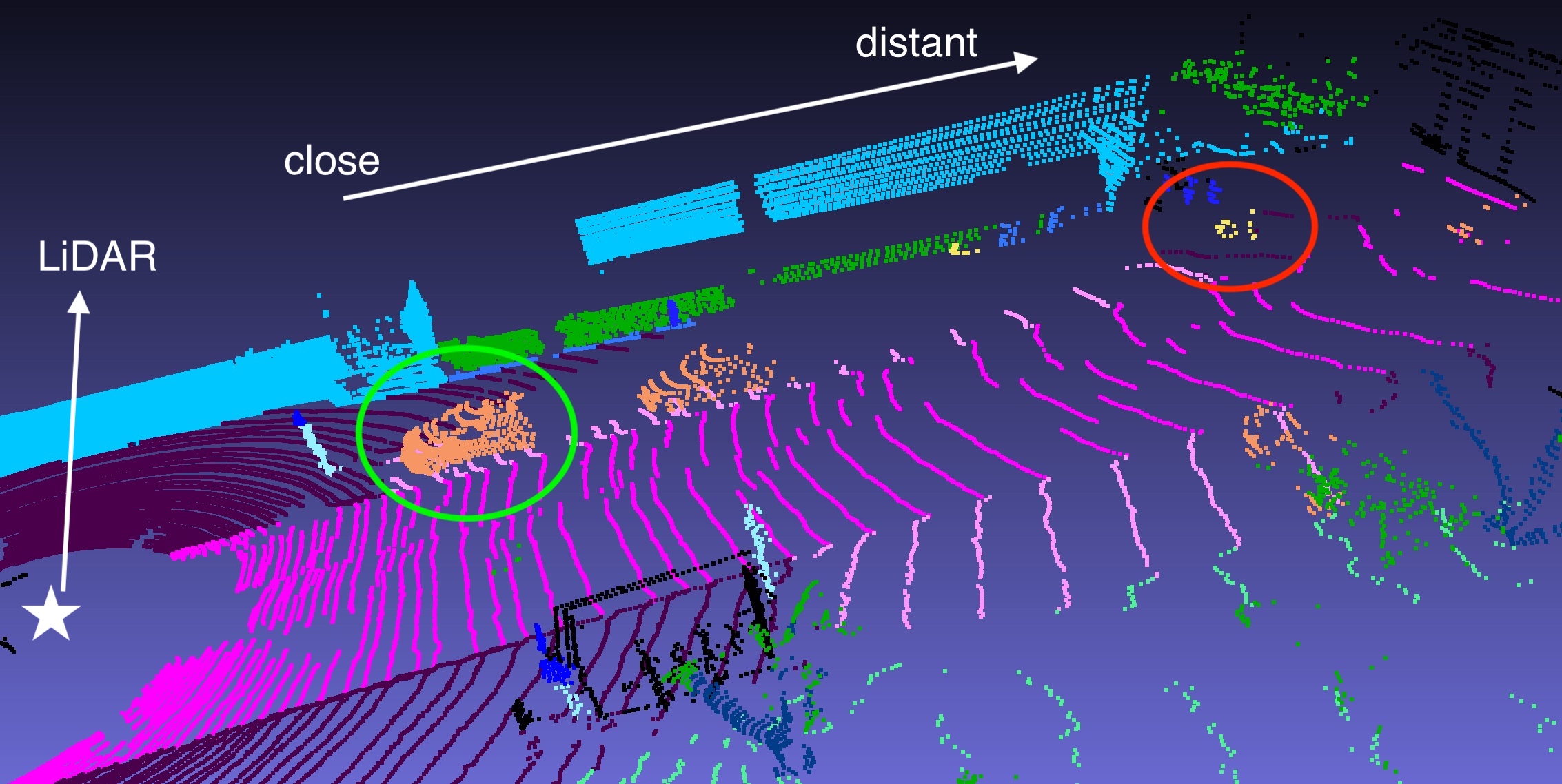}
\end{center}
\vspace{-0.5cm}
\caption{Varying-sparsity property of LiDAR point clouds. The dense close \textit{car} is marked with a green circle and the sparse distant \textit{bicycle} is marked with a red circle (best viewed in color).}
\label{fig:car_and_bicycle}
\vspace{-0.3cm}
\end{figure}

Point clouds scanned by LiDAR have the varying-sparsity property --- close points are dense and distant points are much sparser. This property makes points at different locations perceive different amounts of local information. For example, as shown in Fig.~\ref{fig:car_and_bicycle}, a point of the \textit{car} (circled in green) near the LiDAR is with rich local geometric information from its dense neighbors, which is already enough for the model to make a correct prediction -- incurring more global contexts might be contrarily detrimental. However, a point of \textit{bicycle} (circled in red) far away from the LiDAR lacks shape information due to the extreme sparsity and even occlusion. Then we should supply long-range contexts as a supplement. This example shows treating all the query points equally is not optimal. We thus propose to dynamically select local or global features to address this issue.

\begin{figure}
\begin{center}
\includegraphics[width=1.0\linewidth]{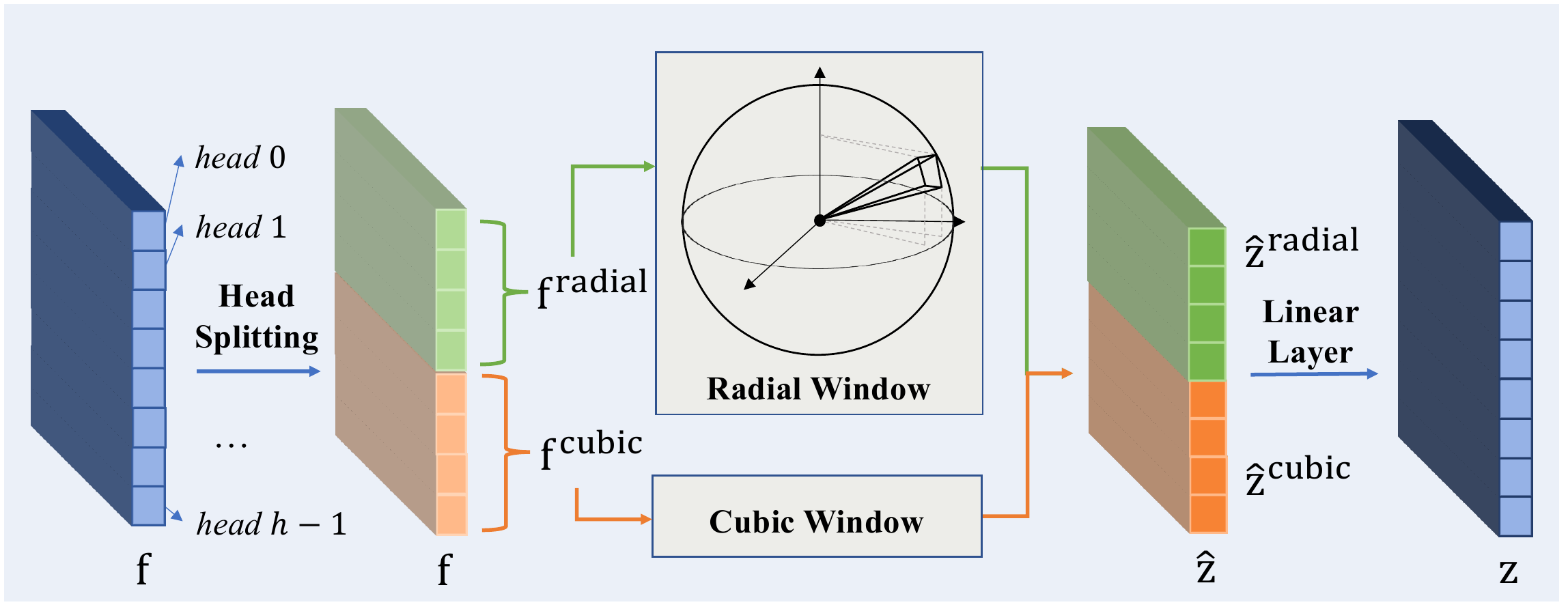}
\end{center}
\vspace{-0.5cm}
\caption{Dynamic feature selection. We split the heads to conduct radial and cubic window self-attention respectively.}
\label{fig:feat_concat}
\vspace{-0.5cm}
\end{figure}

As shown in Fig.~\ref{fig:feat_concat}, for each token, we incorporate not only the radial contextual information, but also local neighbor communication. Specifically, input features are projected into query, key and value features as Eq.~(\ref{eq:qkv_proj}). Then, the first half of the heads are used for radial window self-attention, and the remaining ones are used for cubic window self-attention. After that, these two features are concatenated and then linearly projected to the final output $\mathbf{z}$ for feature fusion. It enables different points to dynamically select local or global features.  Formally, the Equations~(\ref{eq:attn_softmax}-\ref{eq:concat}) are updated to
\begin{equation}
\begin{footnotesize}
\begin{aligned}
    \mathbf{attn}^{radial}_k = \mathbf{softmax}(\mathbf{q}^{radial}_k \cdot {\mathbf{k}^{radial}_k}^T), \nonumber
\end{aligned}
\end{footnotesize}
\end{equation}
\begin{equation}
\begin{footnotesize}
\begin{aligned}
    \hat{\mathbf{z}}^{radial}_{k} = \mathbf{attn}^{radial}_k \cdot {\mathbf{v}^{radial}_k}, \nonumber
\end{aligned}
\end{footnotesize}
\end{equation}
\begin{equation}
\begin{footnotesize}
\begin{aligned}
\mathbf{attn}^{cubic}_k = \mathbf{softmax}(\mathbf{q}^{cubic}_k \cdot {\mathbf{k}^{cubic}_k}^T), \nonumber
\end{aligned}
\end{footnotesize}
\end{equation}
\begin{equation}
\begin{footnotesize}
\begin{aligned}
    \hat{\mathbf{z}}^{cubic}_{k} = \mathbf{attn}^{cubic}_k \cdot {\mathbf{v}^{cubic}_k}, \nonumber
\end{aligned}
\end{footnotesize}
\end{equation}
\begin{equation}
\begin{footnotesize}
\begin{aligned}
    \hat{\mathbf{z}} = \mathbf{concat}(\{\hat{\mathbf{z}}^{radial}_0, \hat{\mathbf{z}}^{radial}_1, ..., \hat{\mathbf{z}}^{radial}_{h/2-1}, \hat{z}^{cubic}_{h/2}, ..., \hat{\mathbf{z}}^{cubic}_{h-1}\}), \nonumber
\end{aligned}
\end{footnotesize}
\end{equation}
\noindent where $\mathbf{q}^{cubic}_k, \mathbf{k}^{cubic}_k, \mathbf{v}^{cubic}_k \in \mathbb{R}^{n^{cubic} \times d}$ denote the query, key and value features for the $k$-th head with cubic window partition, and $\mathbf{attn}^{cubic}_k \in \mathbb{R}^{n^{cubic} \times n^{cubic}}$ denotes the cubic window attention weight for the $k$-th head.



\section{Experiments}

In this section, we first introduce the experimental setting in Sec.~\ref{sec:exp_setting}. Then, we show the semantic segmentation and object detection results in Sec.~\ref{sec:semseg} and \ref{sec:det}. The ablation study and visual comparison are shown in Sec.~\ref{sec:ablation} and \ref{sec:vis_comp}. Our code and models will be made publicly available.

\subsection{Experimental Setting}
\label{sec:exp_setting}
\paragraph{Network Architecture.} For semantic segmentation, we adopt the encoder-decoder structure and follow U-Net~\cite{ronneberger2015u} to concatenate the fine-grained encoder features in the decoder. We follow \cite{zhu2021cylindrical} to use SparseConv~\cite{3DSemanticSegmentationWithSubmanifoldSparseConvNet,SubmanifoldSparseConvNet} as our baseline model. There are a total of 5 stages whose channel numbers are $[32,64,128,256,256]$, and there are two residual blocks at each stage. Our proposed module is stacked at the end of each encoding stage. For object detection, we adopt CenterPoint~\cite{yin2021center} as our baseline model, where the backbone possesses 4 stages whose channel numbers are $[16,32,64,128]$. Our proposed module is stacked at the end of the second and third stages. Note that our proposed module incurs negligible extra parameters, and more details are given in the supplementary material.

\vspace{-0.2cm}
\paragraph{Datasets.} Following previous work, we evaluate methods on nuScenes~\cite{caesar2020nuscenes}, SemanticKITTI~\cite{behley2019semantickitti}, and Waymo Open Dataset~\cite{sun2020scalability} (WOD) for semantic segmentation. For object detection, we evaluate our methods on the nuScenes~\cite{caesar2020nuscenes} dataset. The details of the datasets are given in the supplementary material.

\begin{table*}[!htbp]
    \centering
    \tabcolsep=0.12cm
    {
        \begin{footnotesize}
        \begin{tabular}{ l | c | c c c c c c c c c c c c c c c c c c c}
            \toprule
            Method & mIoU & \rotatebox{90}{road} & \rotatebox{90}{sidewalk} & \rotatebox{90}{parking} & \rotatebox{90}{other-gro.} & \rotatebox{90}{building} & \rotatebox{90}{car} & \rotatebox{90}{truck} & \rotatebox{90}{bicycle} & \rotatebox{90}{motorcycle} & \rotatebox{90}{other-veh.} & \rotatebox{90}{vegetation} & \rotatebox{90}{trunk} & \rotatebox{90}{terrain} & \rotatebox{90}{person} & \rotatebox{90}{bicyclist} & \rotatebox{90}{motorcyclist} & \rotatebox{90}{fence} & \rotatebox{90}{pole} & \rotatebox{90}{traffic sign} \\

            \specialrule{0em}{0pt}{1pt}
            \hline
            \specialrule{0em}{0pt}{1pt}
            
            
            SqueezeSegV2~\cite{wu2019squeezesegv2} & 39.7 & 88.6 & 67.6 & 45.8 & 17.7 & 73.7 & 81.8 & 13.4 & 18.5 & 17.9 & 14.0 & 71.8 & 35.8 & 60.2 & 20.1 & 25.1 & 3.9 & 41.1 & 20.2 & 26.3 \\
            
            DarkNet53Seg~\cite{behley2019semantickitti} & 49.9 & 91.8 & 74.6 & 64.8 & 27.9 & 84.1 & 86.4 & 25.5 & 24.5 & 32.7 & 22.6 & 78.3 & 50.1 & 64.0 & 36.2 & 33.6 & 4.7 & 55.0 & 38.9 & 52.2 \\
            
            RangeNet53++~\cite{milioto2019rangenet++} & 52.2 & 91.8 & 75.2 & 65.0 & 27.8 & 87.4 & 91.4 & 25.7 & 25.7 & 34.4 & 23.0 & 80.5 & 55.1 & 64.6 & 38.3 & 38.8 & 4.8 & 58.6 & 47.9 & 55.9 \\
            
            3D-MiniNet~\cite{alonso20203d} & 55.8 & 91.6 & 74.5 & 64.2 & 25.4 & 89.4 & 90.5 & 28.5 & 42.3 & 42.1 & 29.4 & 82.8 & 60.8 & 66.7 & 47.8 & 44.1 & 14.5 & 60.8 & 48.0 & 56.6 \\
            
            SqueezeSegV3~\cite{xu2020squeezesegv3} & 55.9 & 91.7 & 74.8 & 63.4 & 26.4 & 89.0 & 92.5 & 29.6 & 38.7 & 36.5 & 33.0 & 82.0 & 58.7 & 65.4 & 45.6 & 46.2 & 20.1 & 59.4 & 49.6 & 58.9 \\
            
            PointNet++~\cite{qi2017pointnet++} & 20.1 & 72.0 & 41.8 & 18.7 & 5.6 & 62.3 & 53.7 & 0.9 & 1.9 & 0.2 & 0.2 & 46.5 & 13.8 & 30.0 & 0.9 & 1.0 & 0.0 & 16.9 & 6.0 & 8.9 \\
            
            TangentConv~\cite{tatarchenko2018tangent} & 40.9 & 83.9 & 63.9 & 33.4 & 15.4 & 83.4 & 90.8 & 15.2 & 2.7 & 16.5 & 12.1 & 79.5 & 49.3 & 58.1 & 23.0 & 28.4 & 8.1 & 49.0 & 35.8 & 28.5 \\
            
            PointASNL~\cite{yan2020pointasnl} & 46.8 & 87.4 & 74.3 & 24.3 & 1.8 & 83.1 & 87.9 & 39.0 & 0.0 & 25.1 & 29.2 & 84.1 & 52.2 & 70.6 & 34.2 & 57.6 & 0.0 & 43.9 & 57.8 & 36.9 \\
            
            RandLA-Net~\cite{hu2020randla} & 55.9 & 90.5 & 74.0 & 61.8 & 24.5 & 89.7 & 94.2 & 43.9 & 29.8 & 32.2 & 39.1 & 83.8 & 63.6 & 68.6 & 48.4 & 47.4 & 9.4 & 60.4 & 51.0 & 50.7 \\
            
            KPConv~\cite{thomas2019kpconv} & 58.8 & 90.3 & 72.7 & 61.3 & 31.5 & 90.5 & 95.0 & 33.4 & 30.2 & 42.5 & 44.3 & 84.8 & 69.2 & 69.1 & 61.5 & 61.6 & 11.8 & 64.2 & 56.4 & 47.4 \\
            
            PolarNet~\cite{zhang2020polarnet} & 54.3 & 90.8 & 74.4 & 61.7 & 21.7 & 90.0 & 93.8 & 22.9 & 40.3 & 30.1 & 28.5 & 84.0 & 65.5 & 67.8 & 43.2 & 40.2 & 5.6 & 61.3 & 51.8 & 57.5 \\
            
            JS3C-Net~\cite{yan2021sparse} & 66.0 & 88.9 & 72.1 & 61.9 & 31.9 & 92.5 & 95.8 & 54.3 & 59.3 & 52.9 & 46.0 & 84.5 & 69.8 & 67.9 & 69.5 & 65.4 & 39.9 & 70.8 & 60.7 & 68.7 \\
            
            SPVNAS~\cite{tang2020searching} & 67.0 & 90.2 & 75.4 & 67.6 & 21.8 & 91.6 & 97.2 & 56.6 & 50.6 & 50.4 & 58.0 & 86.1 & 73.4 & 71.0 & 67.4 & 67.1 & 50.3 & 66.9 & 64.3 & 67.3 \\
            
            Cylinder3D~\cite{zhu2021cylindrical} & 68.9 & 92.2 & 77.0 & 65.0 & 32.3 & 90.7 & 97.1 & 50.8 & 67.6 & 63.8 & 58.5 & 85.6 & 72.5 & 69.8 & 73.7 & 69.2 & 48.0 & 66.5 & 62.4 & 66.2 \\
            
            RPVNet~\cite{xu2021rpvnet} & 70.3 & 93.4 & 80.7 & 70.3 & 33.3 & 93.5 & 97.6 & 44.2 & 68.4 & 68.7 & 61.1 & 86.5 & 75.1 & 71.7 & 75.9 & 74.4 & 43.4 & 72.1 & 64.8 & 61.4 \\
            
            (AF)\textsuperscript{2}-S3Net~\cite{cheng20212} & 70.8 & 92.0 & 76.2 & 66.8 & 45.8 & 92.5 & 94.3 & 40.2 & 63.0 & 81.4 & 40.0 & 78.6 & 68.0 & 63.1 & 76.4 & 81.7 & 77.7 & 69.6 & 64.0 & 73.3 \\
            
            PVKD~\cite{hou2022point} & 71.2 & 91.8 & 70.9 & 77.5 & 41.0 & 92.4 & 97.0 & 67.9 & 69.3 & 53.5 & 60.2 & 86.5 & 73.8 & 71.9 & 75.1 & 73.5 & 50.5 & 69.4 & 64.9 & 65.8\\
            
            2DPASS~\cite{yan20222dpass} & 72.9 & 89.7 & 74.7 & 67.4 & 40.0 & 93.5 & 97.0 & 61.1 & 63.6 & 63.4 & 61.5 & 86.2 & 73.9 & 71.0 & 77.9 & 81.3 & 74.1 & 72.9 & 65.0 & 70.4 \\

            \specialrule{0em}{0pt}{1pt}
            \hline
            \specialrule{0em}{0pt}{1pt}
            
            Ours & \textbf{74.8} & 91.8 & 78.2 & 69.7 & 41.3 & 93.8 & 97.5 & 59.6 & 70.1 & 70.5 & 67.7 & 86.7 & 75.1 & 72.4 & 79.0 & 80.4 & 75.3 & 72.8 & 66.8 & 72.9 \\
            
            \bottomrule                                   
        \end{tabular}
        \end{footnotesize}
    }
    \vspace{-0.3cm}
    \caption{Semantic segmentation results on SemanticKITTI \textit{test} set. Methods published before the submission deadline (11/11/2022) are listed.}
    \label{tab:exp_semantickitti}   
\end{table*}

\begin{table*}[!htbp]
    \centering
    \tabcolsep=0.13cm
    {
        \begin{footnotesize}
        \begin{tabular}{ l | c | c c | c c c c c c c c c c c c c c c c}
            \toprule
            
            Method & Input & mIoU & FW mIoU & \rotatebox{90}{barrier} & \rotatebox{90}{bicycle} & \rotatebox{90}{bus} & \rotatebox{90}{car} & \rotatebox{90}{construction} & \rotatebox{90}{motorcycle} & \rotatebox{90}{pedestrian} & \rotatebox{90}{traffic cone} & \rotatebox{90}{trailer} & \rotatebox{90}{truck} & \rotatebox{90}{driveable} & \rotatebox{90}{other flat} & \rotatebox{90}{sidewalk} & \rotatebox{90}{terrain} & \rotatebox{90}{manmade} & \rotatebox{90}{vegetation} \\
            
            \specialrule{0em}{0pt}{1pt}
            \hline
            \specialrule{0em}{0pt}{1pt}
            
            PolarNet~\cite{zhang2020polarnet} & L & 69.4 & 87.4 & 72.2 & 16.8 & 77.0 & 86.5 & 51.1 & 69.7 & 64.8 & 54.1 & 69.7 & 63.5 & 96.6 & 67.1 & 77.7 & 72.1 & 87.1 & 84.5 \\
            
            JS3C-Net~\cite{yan2021sparse} & L & 73.6 & 88.1 & 80.1 & 26.2 & 87.8 & 84.5 & 55.2 & 72.6 & 71.3 & 66.3 & 76.8 & 71.2 & 96.8 & 64.5 & 76.9 & 74.1 & 87.5 & 86.1 \\
            
            Cylinder3D~\cite{zhu2021cylindrical} & L & 77.2 & 89.9 & 82.8 & 29.8 & 84.3 & 89.4 & 63.0 & 79.3 & 77.2 & 73.4 & 84.6 & 69.1 & 97.7 & 70.2 & 80.3 & 75.5 & 90.4 & 87.6 \\
            
            AMVNet~\cite{liong2020amvnet} & L & 77.3 & 90.1 & 80.6 & 32.0 & 81.7 & 88.9 & 67.1 & 84.3 & 76.1 & 73.5 & 84.9 & 67.3 & 97.5 & 67.4 & 79.4 & 75.5 & 91.5 & 88.7 \\
            
            SPVCNN~\cite{tang2020searching} & L & 77.4 & 89.7 & 80.0 & 30.0 & 91.9 & 90.8 & 64.7 & 79.0 & 75.6 & 70.9 & 81.0 & 74.6 & 97.4 & 69.2 & 80.0 & 76.1 & 89.3 & 87.1 \\
            
            (AF)\textsuperscript{2}-S3Net~\cite{cheng20212} & L & 78.3 & 88.5 & 78.9 & 52.2 & 89.9 & 84.2 & 77.4 & 74.3 & 77.3 & 72.0 & 83.9 & 73.8 & 97.1 & 66.5 & 77.5 & 74.0 & 87.7 & 86.8 \\
            
            PMF~\cite{zhuang2021perception} & L+C & 77.0 & 89.0 & 82.0 & 40.0 & 81.0 & 88.0 & 64.0 & 79.0 & 80.0 & 76.0 & 81.0 & 67.0 & 97.0 & 68.0 & 78.0 & 74.0 & 90.0 & 88.0 \\
            
            2D3DNet~\cite{genova2021learning} & L+C & 80.0 & 90.1 & 83.0 & 59.4 & 88.0 & 85.1 & 63.7 & 84.4 & 82.0 & 76.0 & 84.8 & 71.9 & 96.9 & 67.4 & 79.8 & 76.0 & 92.1 & 89.2 \\
            
            2DPASS~\cite{yan20222dpass} & L & 80.8 & 90.1 & 81.7 & 55.3 & 92.0 & 91.8 & 73.3 & 86.5 & 78.5 & 72.5 & 84.7 & 75.5 & 97.6 & 69.1 & 79.9 & 75.5 & 90.2 & 88.0 \\
            
            \specialrule{0em}{0pt}{1pt}
            \hline
            \specialrule{0em}{0pt}{1pt}
            
            Ours & L & \textbf{81.9} & \textbf{91.7} & 83.3 & 39.2 & 94.7 & 92.5 & 77.5 & 84.2 & 84.4 & 79.1 & 88.4 & 78.3 & 97.9 & 69.0 & 81.5 & 77.2 & 93.4 & 90.2 \\
            
            \bottomrule                                   
        \end{tabular}
        \end{footnotesize}
    }
    \vspace{-0.3cm}
    \caption{Semantic segmentation results on nuScenes \textit{test} set. Methods published before the submission deadline (11/11/2022) are listed.}
    \label{tab:exp_nuscenes}   
\vspace{-0.2cm}
\end{table*}

\vspace{-0.2cm}
\paragraph{Implementation Details.} For semantic segmentation, we use 4 GeForce RTX 3090 GPUs for training. We train the models for 50 epochs with AdamW~\cite{loshchilov2017decoupled} optimizer and `poly' scheduler where \textit{power} is set to 0.9. The learning rate and weight decay are set to $0.006$ and $0.01$, respectively. Batch size is set to 16 on nuScenes, and 8 on both SemanticKITTI and Waymo Open Dataset. The window size is set to $[120m,2^{\circ},2^{\circ}]$ for $(r, \theta, \phi)$ on both nuScenes and SemanticKITTI, and $[80m,1.5^{\circ},1.5^{\circ}]$ on Waymo Open Dataset. During data preprocessing, we confine the input scene to the range from $[-51.2m, -51.2m, -4m]$ to $[51.2m, 51.2m, 2.4m]$ on SemanticKITTI and $[-75.2m, -75.2m, -2m]$ to $[75.2m, 75.2m, 4m]$ on Waymo. Also, we set the voxel size to $0.1m$ on both nuScenes and Waymo, and $0.05m$ on SemanticKITTI.

For object detection, we adopt the OpenPCDet~\cite{openpcdet2020} codebase and follow the default CenterPoint~\cite{yin2021center} to set the training hyper-parameters. We set the window size to $[120m,1.5^{\circ},1.5^{\circ}]$.

\subsection{Semantic Segmentation Results}
\label{sec:semseg}


\begin{table*}[!htbp]
    \centering
    \tabcolsep=0.19cm
    {
        \begin{footnotesize}
        \begin{tabular}{ l | c | c c c c c c c c c c c c c c c c}
            \toprule
            
            Method & mIoU & \rotatebox{90}{barrier} & \rotatebox{90}{bicycle} & \rotatebox{90}{bus} & \rotatebox{90}{car} & \rotatebox{90}{construction} & \rotatebox{90}{motorcycle} & \rotatebox{90}{pedestrian} & \rotatebox{90}{traffic cone} & \rotatebox{90}{trailer} & \rotatebox{90}{truck} & \rotatebox{90}{driveable} & \rotatebox{90}{other flat} & \rotatebox{90}{sidewalk} & \rotatebox{90}{terrain} & \rotatebox{90}{manmade} & \rotatebox{90}{vegetation} \\
            
            \specialrule{0em}{0pt}{1pt}
            \hline
            \specialrule{0em}{0pt}{1pt}
            
            RangeNet53++~\cite{milioto2019rangenet++} & 65.5 & 66.0 & 21.3 & 77.2 & 80.9 & 30.2 & 66.8 & 69.6 & 52.1 & 54.2 & 72.3 & 94.1 & 66.6 & 63.5 & 70.1 & 83.1 & 79.8 \\
            
            PolarNet~\cite{zhang2020polarnet} & 71.0 & 74.7 & 28.2 & 85.3 & 90.9 & 35.1 & 77.5 & 71.3 & 58.8 & 57.4 & 76.1 & 96.5 & 71.1 & 74.7 & 74.0 & 87.3 & 85.7 \\
            
            Salsanext~\cite{cortinhal2020salsanext} & 72.2 & 74.8 & 34.1 & 85.9 & 88.4 & 42.2 & 72.4 & 72.2 & 63.1 & 61.3 & 76.5 & 96.0 & 70.8 & 71.2 & 71.5 & 86.7 & 84.4 \\
            
            AMVNet~\cite{liong2020amvnet} & 76.1 & 79.8 & 32.4 & 82.2 & 86.4 & 62.5 & 81.9 & 75.3 & 72.3 & 83.5 & 65.1 & 97.4 & 67.0 & 78.8 & 74.6 & 90.8 & 87.9 \\
            
            Cylinder3D~\cite{zhu2021cylindrical} & 76.1 & 76.4 & 40.3 & 91.2 & 93.8 & 51.3 & 78.0 & 78.9 & 64.9 & 62.1 & 84.4 & 96.8 & 71.6 & 76.4 & 75.4 & 90.5 & 87.4 \\
            
            PVKD~\cite{hou2022point} & 76.0 & 76.2 & 40.0 & 90.2 & 94.0 & 50.9 & 77.4 & 78.8 & 64.7 & 62.0 & 84.1 & 96.6 & 71.4 & 76.4 & 76.3 & 90.3 & 86.9 \\
            
            RPVNet~\cite{xu2021rpvnet} & 77.6 & 78.2 & 43.4 & 92.7 & 93.2 & 49.0 & 85.7 & 80.5 & 66.0 & 66.9 & 84.0 & 96.9 & 73.5 & 75.9 & 76.0 & 90.6 & 88.9 \\
            
            \specialrule{0em}{0pt}{1pt}
            \hline
            \specialrule{0em}{0pt}{1pt}
            
            Ours & 78.4 & 77.7 & 43.8 & 94.5 & 93.1 & 52.4 & 86.9 & 81.2 & 65.4 & 73.4 & 85.3 & 97.0 & 73.4 & 75.4 & 75.0 & 91.0 & 89.2 \\
            
            Ours$^\ddagger$ & \textbf{79.5} & 78.7 & 46.7 & 95.2 & 93.7 & 54.0 & 88.9 & 81.1 & 68.0 & 74.2 & 86.2 & 97.2 & 74.3 & 76.3 & 75.8 & 91.4 & 89.7 \\
            
            \bottomrule
        \end{tabular}
        \end{footnotesize}
    }
    \vspace{-0.3cm}
    \caption{Semantic segmentation results on nuScenes \textit{val set}. $\;\;^{\ddagger}$ denotes using rotation and translation testing-time augmentations.}
    \label{tab:exp_nuscenes_val}   
\end{table*}

\begin{table*}[!htbp]
    \centering
    \tabcolsep=0.05cm
    {
        \begin{footnotesize}
        \begin{tabular}{ l | c | c c c | c c c c c c c c c c c c c c c c c c c c c c}
            \toprule
            
            Method & mIoU & close & med. & far & \rotatebox{90}{car} & \rotatebox{90}{truck} & \rotatebox{90}{bus} & \rotatebox{90}{other-veh.} & \rotatebox{90}{motorcyclist} & \rotatebox{90}{bicyclist} & \rotatebox{90}{pedestrian} & \rotatebox{90}{sign} & \rotatebox{90}{traffic-light} & \rotatebox{90}{pole} & \rotatebox{90}{con.cone} & \rotatebox{90}{bicycle} & \rotatebox{90}{motorcycle} & \rotatebox{90}{building} & \rotatebox{90}{vegetation} & \rotatebox{90}{tree-trunk} & \rotatebox{90}{curb} & \rotatebox{90}{road} & \rotatebox{90}{lane-marker} & \rotatebox{90}{other-gro.} & \rotatebox{90}{walkable} & \rotatebox{90}{sidewalk}\\
            
            \specialrule{0em}{0pt}{1pt}
            \hline
            \specialrule{0em}{0pt}{1pt}
            
            SparseConv~\cite{SubmanifoldSparseConvNet} & 66.6 & 67.8 & 64.1 & 52.6 & 94.4 & 59.8 & 85.1 & 37.8 & 2.2 & 69.1 & 89.3 & 73.4 & 40.4 & 74.8 & 57.3 & 66.6 & 75.2 & 95.5 & 91.3 & 67.0 & 68.1 & 92.3 & 41.7 & 30.1 & 79.0 & 75.6 \\
            
            \specialrule{0em}{0pt}{1pt}
            \hline
            \specialrule{0em}{0pt}{1pt}
            
            Ours & \textbf{69.9} & 70.3 & 68.6 & 61.9 & 94.5 & 61.6 & 87.7 & 40.2 & 0.9 & 69.7 & 90.2 & 73.9 & 41.8 & 77.2 & 65.4 & 71.9 & 83.7 & 95.9 & 91.7 & 68.4 & 69.8 & 93.3 & 53.9 & 47.9 & 80.8 & 77.2 \\
            
            \bottomrule                                   
        \end{tabular}
        \end{footnotesize}
    }
    \vspace{-0.2cm}
    \caption{Semantic segmentation results on Waymo Open Dataset \textit{val set}.}
    \label{tab:exp_waymo_val}   
\vspace{-0.3cm}
\end{table*}

The results on SemanticKITTI \textit{test} set are shown in Table~\ref{tab:exp_semantickitti}. Our method yields $74.8\%$ mIoU, a new state-of-the-art result. Compared to the methods based on range images~\cite{wu2019squeezesegv2,milioto2019rangenet++} and Bird-Eye-View (BEV)~\cite{zhang2020polarnet}, ours gives a result with over $20\%$ mIoU performance gain. Moreover, thanks to the capability of directly aggregating long-range information, our method significantly outperforms the models based on sparse convolution~\cite{yan2021sparse,zhu2021cylindrical,tang2020searching,cheng20212,xu2021rpvnet}. It is also notable that our method outperforms 2DPASS~\cite{yan20222dpass} that uses extra 2D images in training by $1.9\%$ mIoU.

In Tables~\ref{tab:exp_nuscenes} and \ref{tab:exp_nuscenes_val}, we also show the semantic segmentation results on nuScenes \textit{test} and \textit{val} set, respectively. Our method consistently outperforms others by a large margin, and achieves the 1\textsuperscript{st} place on the benchmark. It is intriguing to note that our method is purely based on LiDAR data, and it works even better than approaches of~\cite{zhuang2021perception,genova2021learning,yan20222dpass} that use additional 2D information.

Moreover, we demonstrate the semantic segmentation results on Waymo Open Dataset \textit{val} set in Table~\ref{tab:exp_waymo_val}. Our model outperforms the baseline model with a substantial gap of $3.3\%$ mIoU. Also, it is worth noting that our method achieves a $9.3\%$ mIoU performance gain for the \textit{far} points, \ie, the sparse distant points.

\subsection{Object Detection Results}
\label{sec:det}

\begin{table}[!t]
    \centering
    \tabcolsep=0.07cm
    {
        \begin{footnotesize}
        \begin{tabular}{ c |  c  c  c | c c c | c c }
            \toprule
            ID & RadialWin & ExpSplit & Dynamic & close & medium & far & overall & $\Delta$ \\

            \specialrule{0em}{0pt}{1pt}
            \hline
            \specialrule{0em}{0pt}{1pt}
            
            \uppercase\expandafter{\romannumeral1} & & & & 78.79 & 51.54 & 13.28 & 75.21 & 0.00 \\ 
            
            \uppercase\expandafter{\romannumeral2} & \Checkmark & & & 78.95 & 57.21 & 26.67 & 76.31 & +1.10 \\ 
            
            \uppercase\expandafter{\romannumeral3} & \Checkmark &  \Checkmark & & 79.92 & 61.09 & 31.10 & 77.60 & +2.39 \\ 
            
            \uppercase\expandafter{\romannumeral4} & \Checkmark &   & \Checkmark & 79.51 & 58.94 & 28.95 & 77.05 & +1.84\\ 
            
            \uppercase\expandafter{\romannumeral5} & \Checkmark &  \Checkmark & \Checkmark & 80.80 & 60.78 & 30.38 & \textbf{78.41} & +3.20\\
            
            \bottomrule                                   
        \end{tabular}
        \end{footnotesize}
    }    
    \vspace{-0.2cm}
    \caption{Ablation study. \textbf{RadialWin}: Radial window shape. \textbf{ExpSplit}: Exponential splitting. \textbf{Dynamic}: Dynamic Feature Selection. Metric: mIoU. }
    \label{tab:ablation}   
\vspace{-0.2cm}
\end{table}

\begin{table}[!t]
    \centering
    \tabcolsep=0.45cm
    {
        \begin{footnotesize}
        \begin{tabular}{ c | c c c | c }
            \toprule
            Method & close & medium & far & overall \\

            \specialrule{0em}{0pt}{1pt}
            \hline
            \specialrule{0em}{0pt}{1pt}
            
            Cubic & 79.21 & 54.31 & 19.31 & 76.19 \\
            
            Radial & 80.80 & 60.78 & 30.38 & 78.41 \\
            
            \bottomrule                                   
        \end{tabular}
        \end{footnotesize}
    }    
    \vspace{-0.2cm}
    \caption{Comparison between radial and cubic window shapes.}
    \label{tab:comp_radial_cubic}   
\vspace{-0.3cm}
\end{table}

\begin{table}[!t]
    \centering
    \tabcolsep=0.45cm
    {
        \begin{footnotesize}
        \begin{tabular}{ c | c c c c }
            \toprule
            window size & $1.0^\circ$ & $1.5^\circ$ & $2.0^\circ$ & $2.5^\circ$ \\

            \specialrule{0em}{0pt}{1pt}
            \hline
            \specialrule{0em}{0pt}{1pt}
            
            mIoU (\%) & 77.8 & 77.5 & 78.4 & 77.6 \\
            
            \bottomrule                                   
        \end{tabular}
        \end{footnotesize}
    }    
    \vspace{-0.2cm}
    \caption{Effect of window size for the $\theta$ and $\phi$ dimensions.}
    \label{tab:exp_window_size}   
\vspace{-0.5cm}
\end{table}

\begin{table*}[!htbp]
    \centering
    \tabcolsep=0.35cm
    {
        \begin{threeparttable}
        \begin{footnotesize}
        \begin{tabular}{ l | c | c | c c c c c c c c c c }
            \toprule
            
            Method & NDS & mAP & Car & Truck & Bus & Trailer & C.V. & Ped. & Mot. & Byc. & T.C. & Bar. \\
            
            \specialrule{0em}{0pt}{1pt}
            \hline
            \specialrule{0em}{0pt}{1pt}
            
            PointPillars~\cite{lang2019pointpillars} & 45.3 & 30.5 & 68.4 & 23.0 & 28.2 & 23.4 & 4.1 & 59.7 & 27.4 & 1.1 & 30.8 & 38.9 \\
            
            3DSSD~\cite{yang20203dssd} & 56.4 & 42.6 & 81.2 & 47.2 & 61.4 & 30.5 & 12.6 & 70.2 & 36.0 & 8.6 & 31.1 & 47.9 \\
            
            CBSG~\cite{zhu2019class} & 63.3 & 52.8 & 81.1 & 48.5 & 54.9 & 42.9 & 10.5 & 80.1 & 51.5 & 22.3 & 70.9 & 65.7 \\
            
            CenterPoint~\cite{yin2021center} & 65.5 & 58.0 & 84.6 & 51.0 & 60.2 & 53.2 & 17.5 & 83.4 & 53.7 & 28.7 & 76.7 & 70.9\\
            
            HotSpotNet~\cite{chen2020object} & 66.0 & 59.3 & 83.1 & 50.9 & 56.4 & 53.3 & 23.0 & 81.3 & 63.5 & 36.6 & 73.0 & 71.6\\
            
            CVCNET~\cite{chen2020every} & 66.6 & 58.2 & 82.6 & 49.5 & 59.4 & 51.1 & 16.2 & 83.0 & 61.8 & 38.8 & 69.7 & 69.7\\
            
            TransFusion~\cite{bai2022transfusion} & 70.2 & 65.5 & 86.2 & 56.7 & 66.3 & 58.8 & 28.2 & 86.1 & 68.3 & 44.2 & 82.0 & 78.2\\
            
            Focals Conv~\cite{chen2022focal} & 70.0 & 63.8 & 86.7 & 56.3 & 67.7 & 59.5 & 23.8 & 87.5 & 64.5 & 36.3 & 81.4 & 74.1\\
            
            \specialrule{0em}{0pt}{1pt}
            \hline
            \specialrule{0em}{0pt}{1pt}
            
            Ours & 70.7 & 65.5 & 84.9 & 55.1 & 66.4 & 59.3 & 29.9 & 86.0 & 71.4 & 47.1 & 79.7 & 75.2 \\
            
            Ours$^{\ddagger}$ &	\textbf{72.8} & \textbf{68.5} & 85.3 & 57.9 & 67.0 & 59.9 & 33.7 & 88.6 & 76.3 & 56.4 & 82.2 & 78.2 \\
            
            \bottomrule                                   
        \end{tabular}
        \end{footnotesize}
        \begin{tablenotes}
          \small
          \item $\;\;^{\ddagger}$ Flipping and rotation testing-time augmentations.
        \end{tablenotes}
        \end{threeparttable}
    }
    \vspace{-0.2cm}
    \caption{Object detection results on nuScenes \textit{test set}. Methods published before the submission deadline (11/11/2022) are listed.}
    \label{tab:exp_nuscenes_det}   
\end{table*}

Our method also achieves strong performance in object detection. As shown in Table~\ref{tab:exp_nuscenes_det}, our method outperforms other published methods on nuScenes \textit{test set}, and ranks 3\textsuperscript{rd} on the LiDAR-only benchmark. It shows that directly aggregating long-range information is also beneficial for object detection. It also manifests the capability of our method to generalize to instance-level tasks.

\begin{figure*}[ht]
	\centering
    \begin{minipage}  {0.16\linewidth}
        \centering
        \includegraphics [width=1\linewidth,height=0.5\linewidth]
        {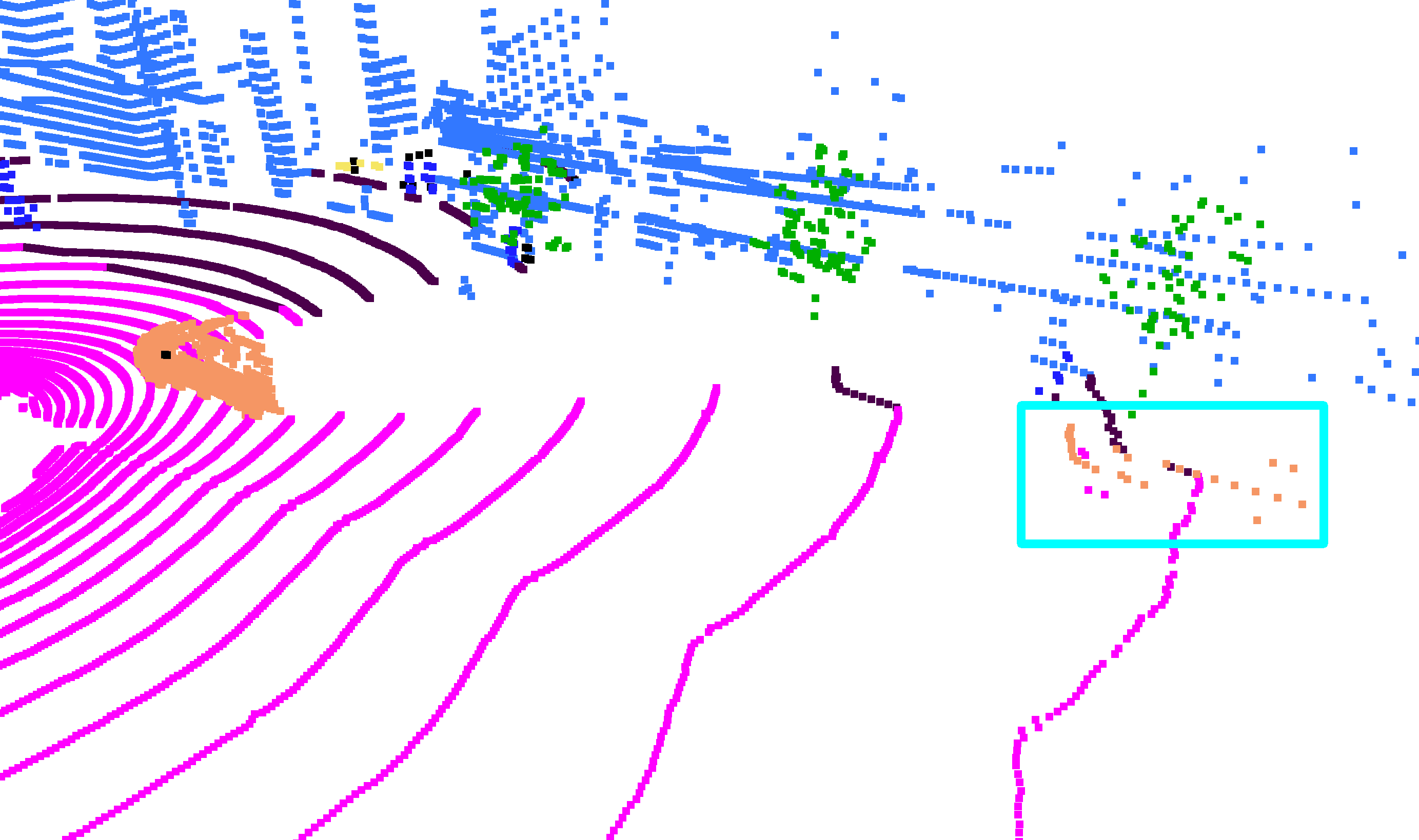}
    \end{minipage}      
    \begin{minipage}  {0.16\linewidth}
        \centering
        \includegraphics [width=1\linewidth,height=0.5\linewidth]
        {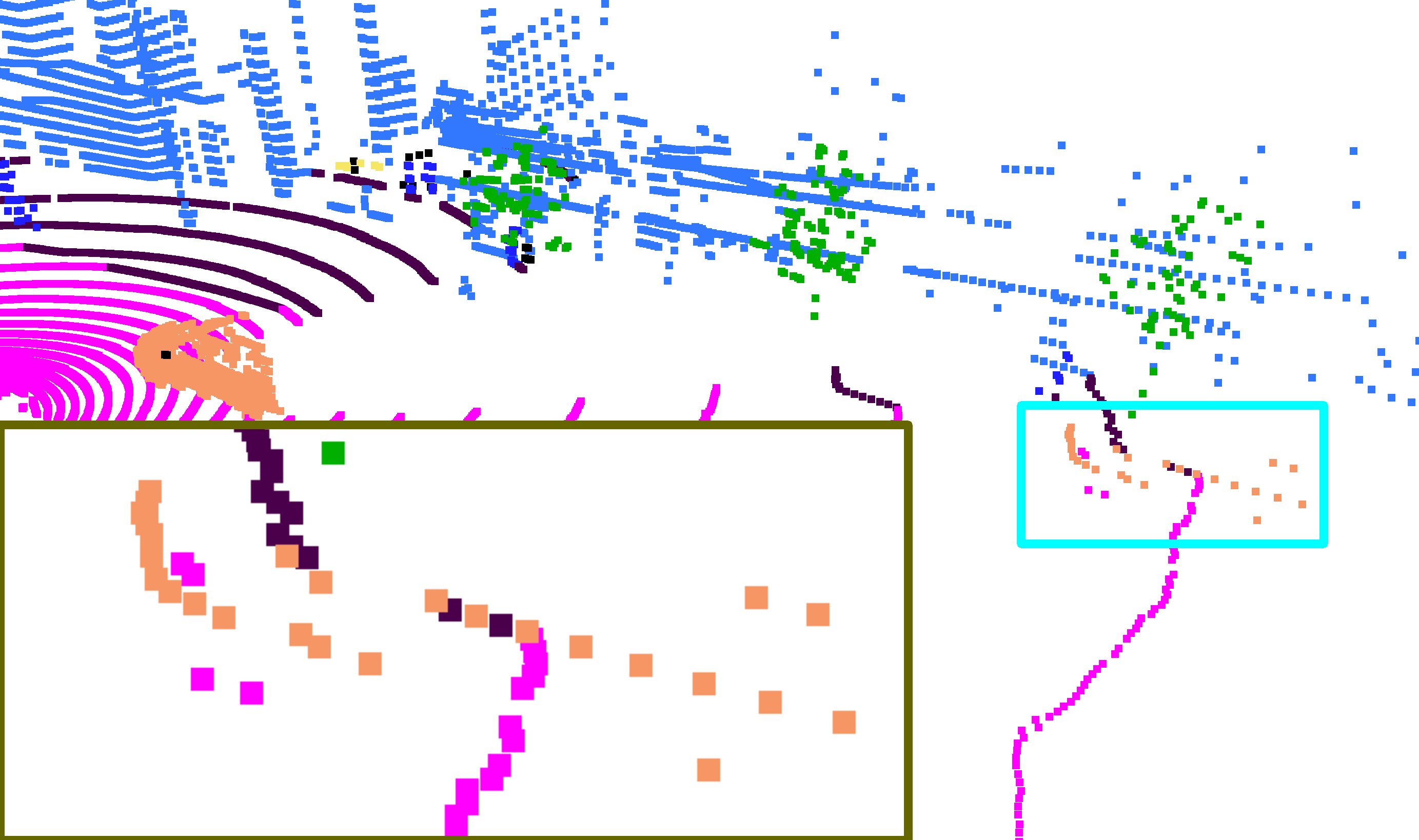}
    \end{minipage}      
     \begin{minipage}  {0.16\linewidth}
        \centering
        \includegraphics [width=1\linewidth,height=0.5\linewidth]
        {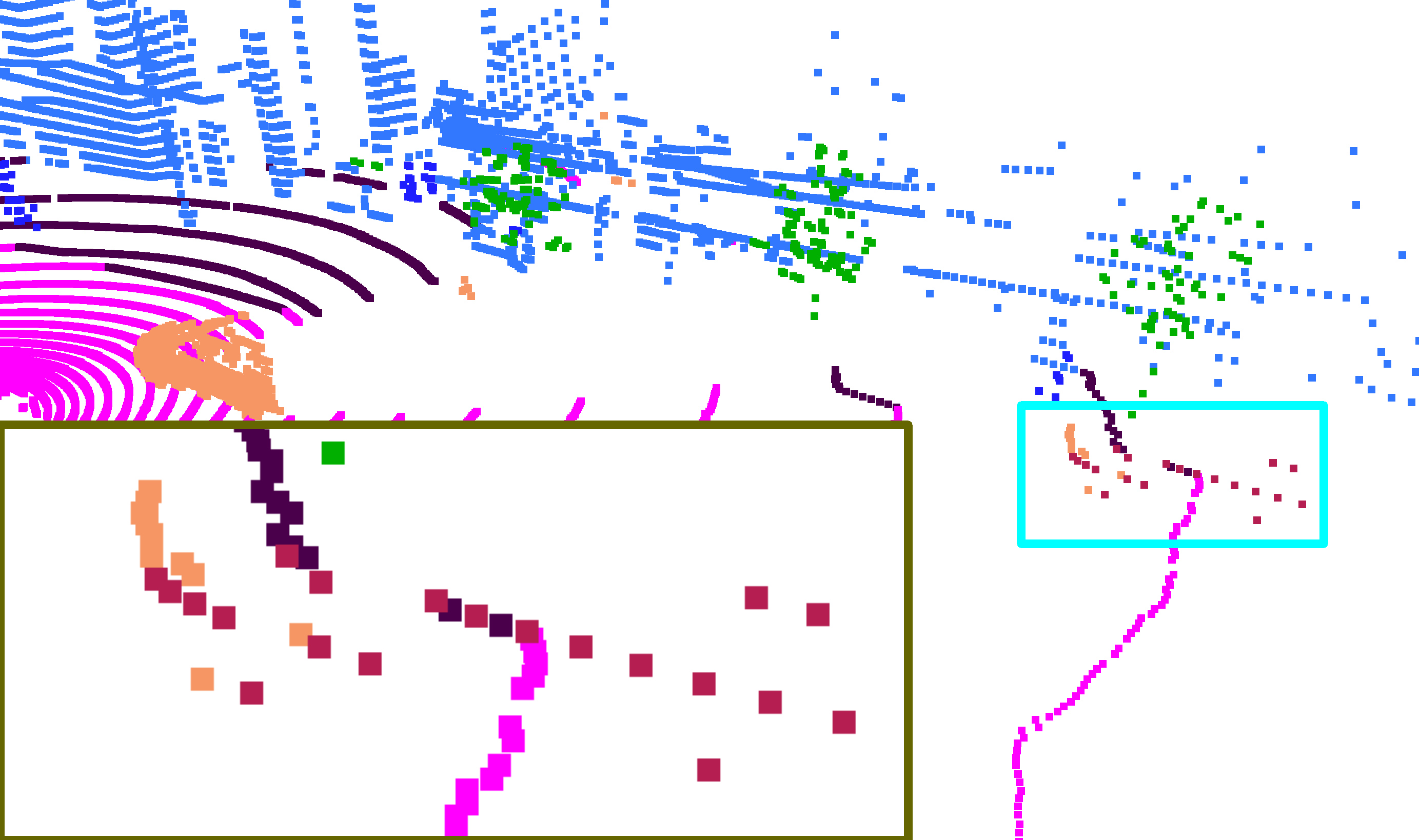}
    \end{minipage} 
    \begin{minipage}  {0.16\linewidth}
        \centering
        \includegraphics [width=1\linewidth,height=0.5\linewidth]
        {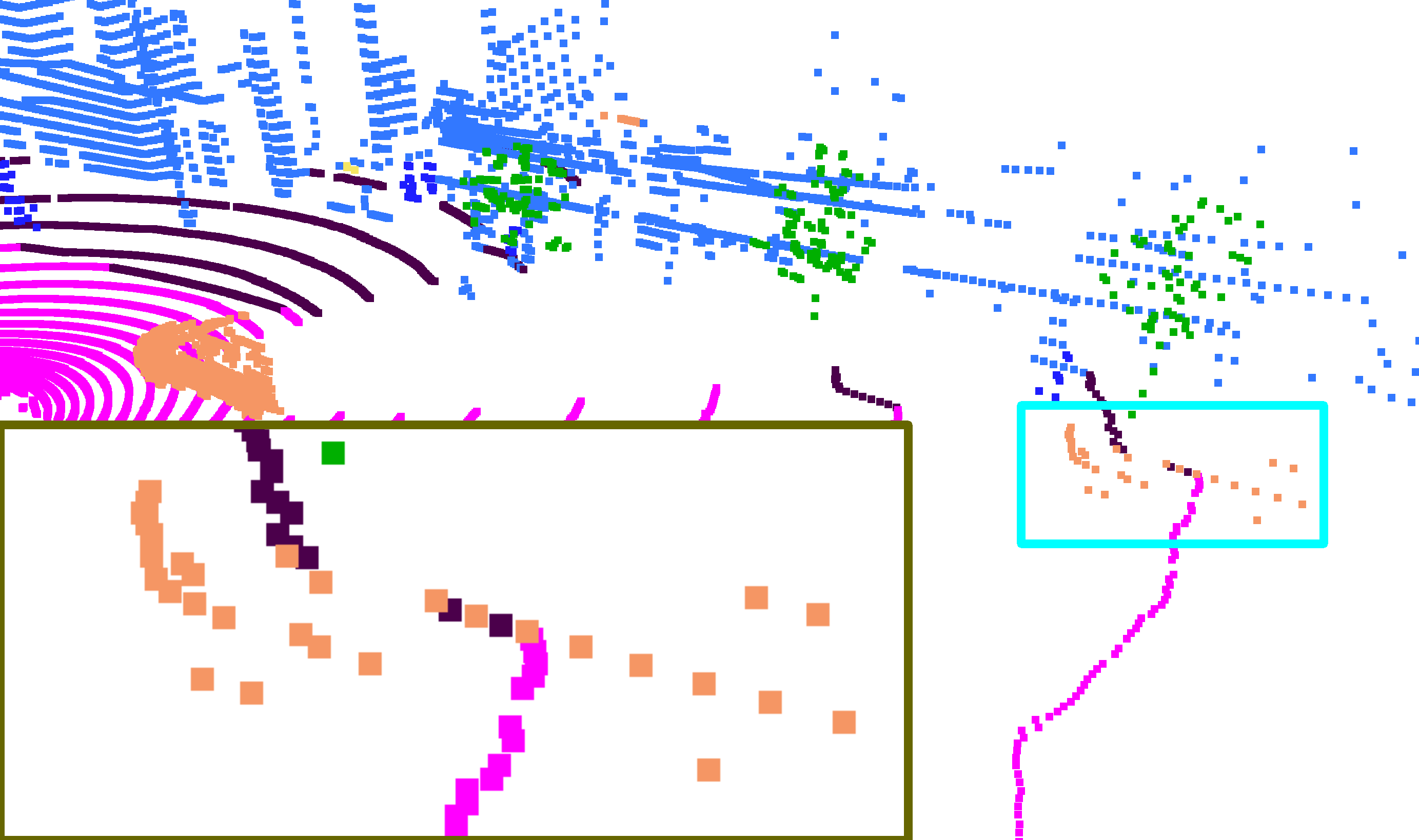}
    \end{minipage}      
     \begin{minipage}  {0.16\linewidth}
        \centering
        \includegraphics [width=1\linewidth,height=0.5\linewidth]
        {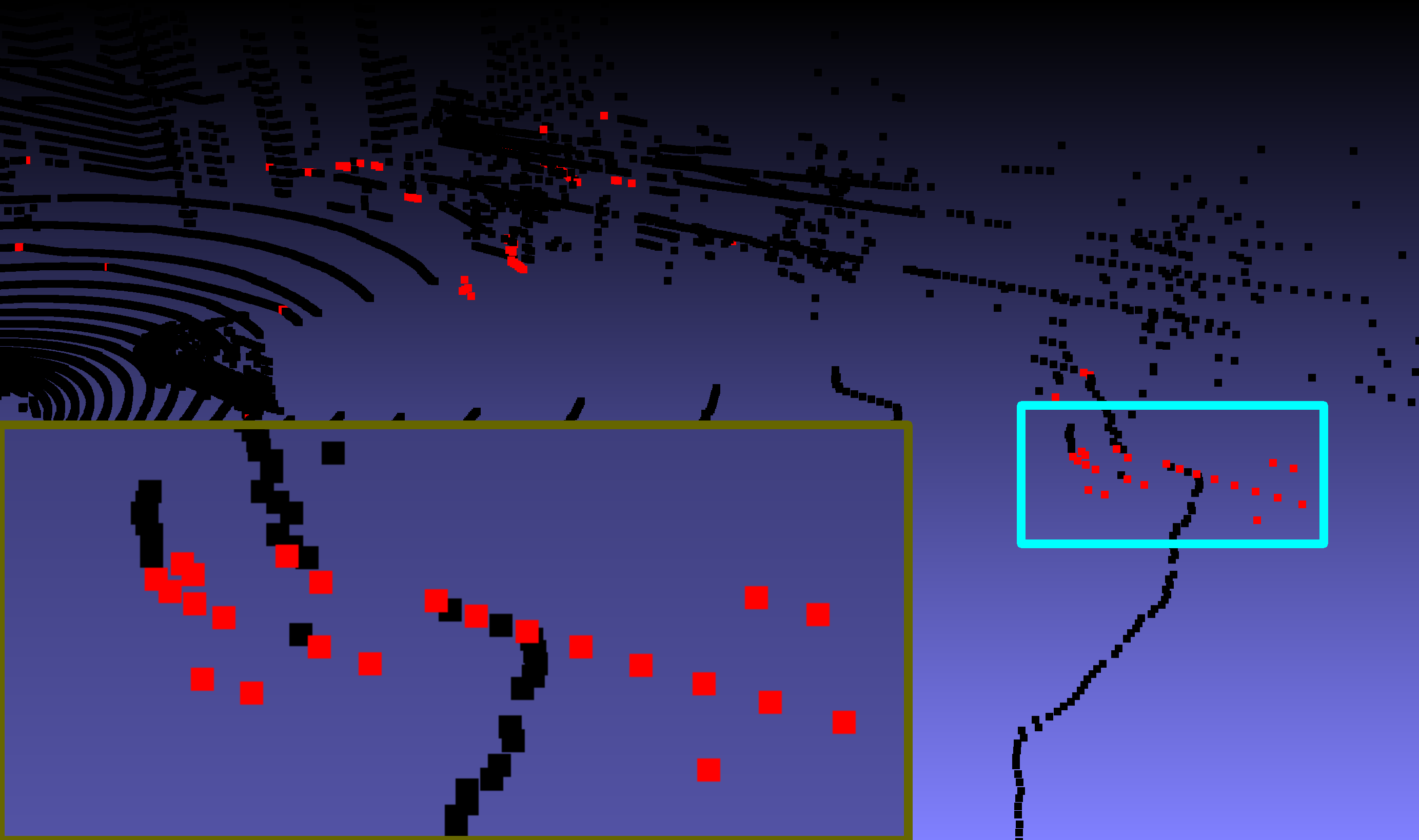}
    \end{minipage} 
     \begin{minipage}  {0.16\linewidth}
        \centering
        \includegraphics [width=1\linewidth,height=0.5\linewidth]
        {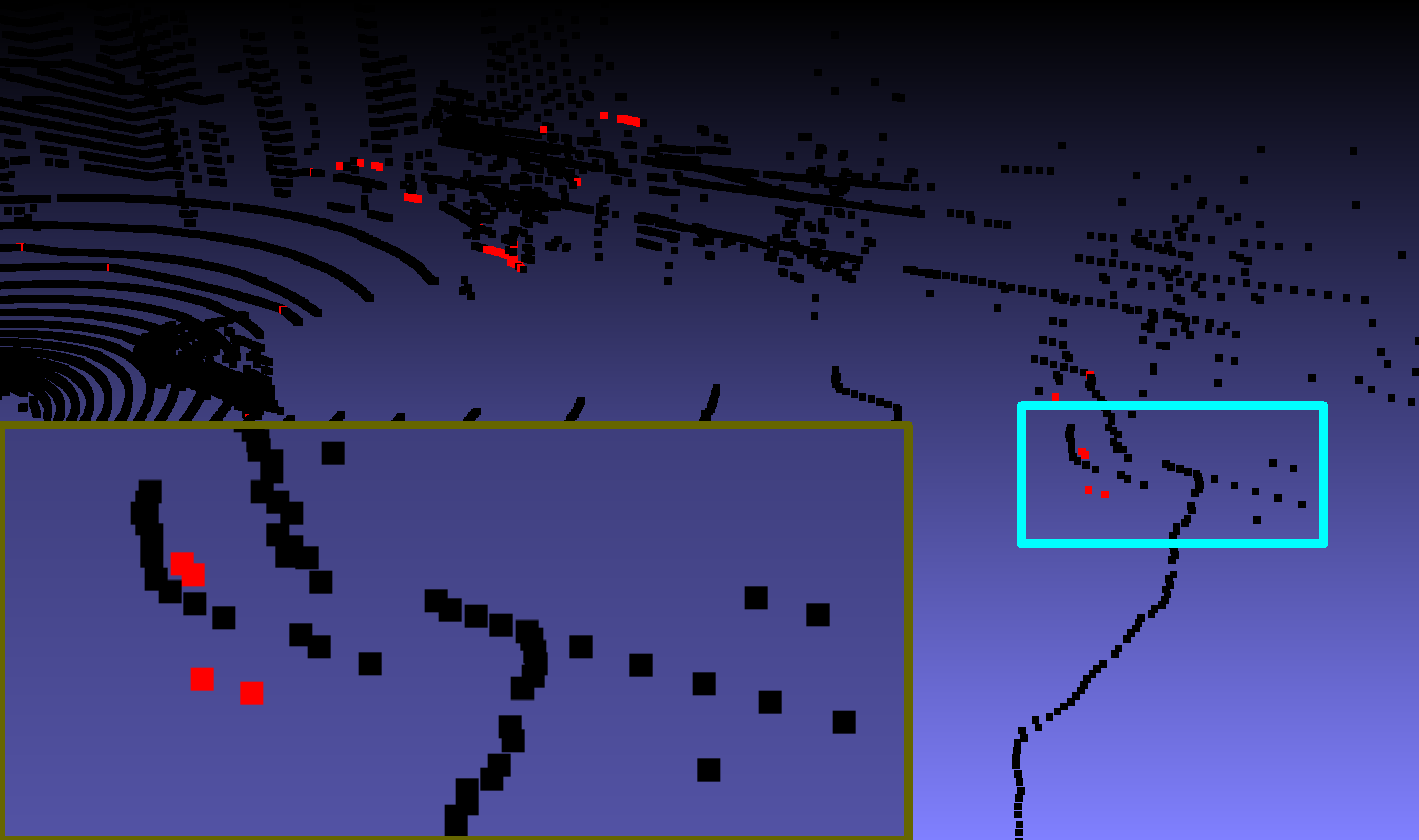}
    \end{minipage} 
	 
    \begin{minipage}  {0.16\linewidth}
        \centering
        \includegraphics [width=1\linewidth,height=0.5\linewidth]
        {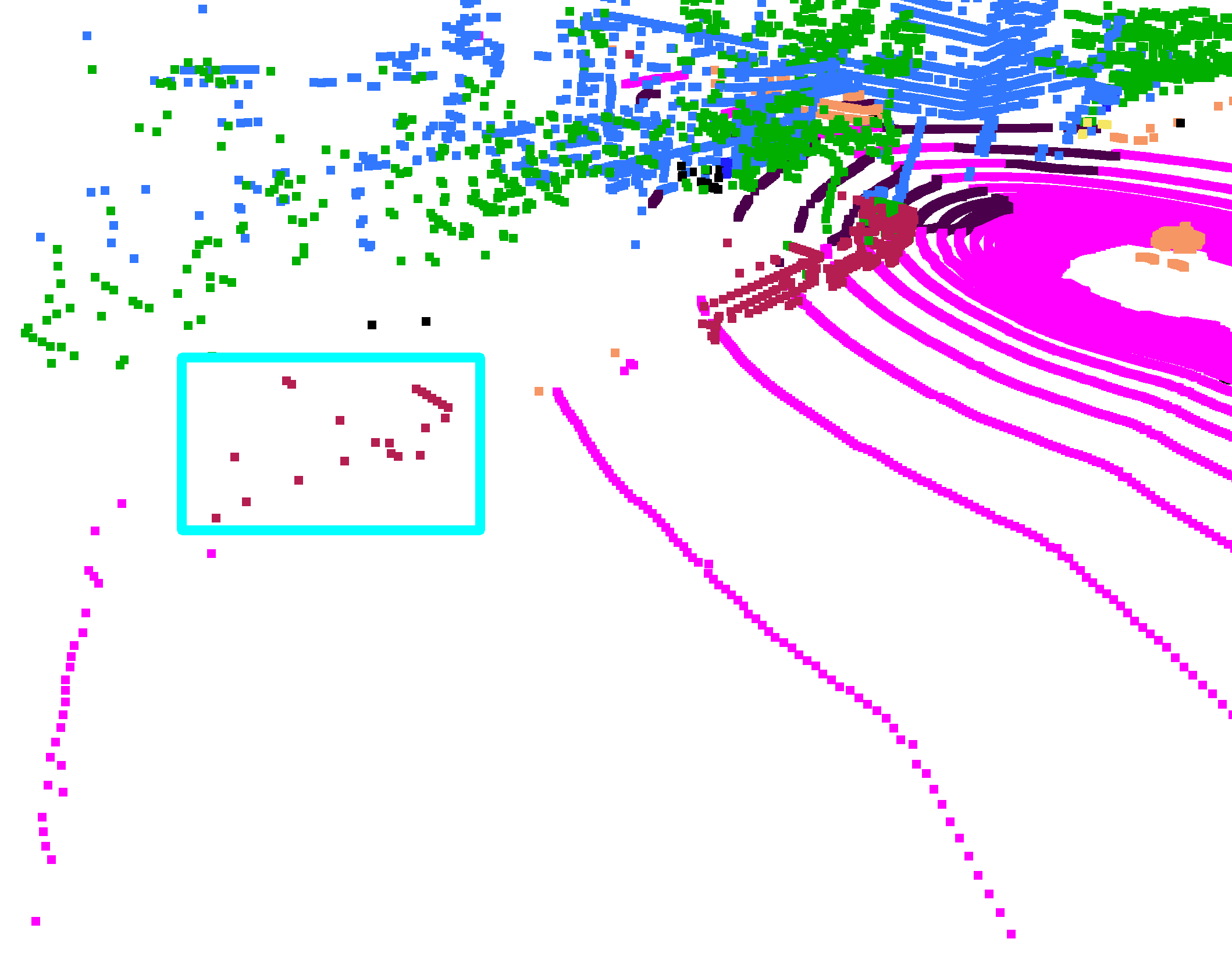}
    \end{minipage}      
    \begin{minipage}  {0.16\linewidth}
        \centering
        \includegraphics [width=1\linewidth,height=0.5\linewidth]
        {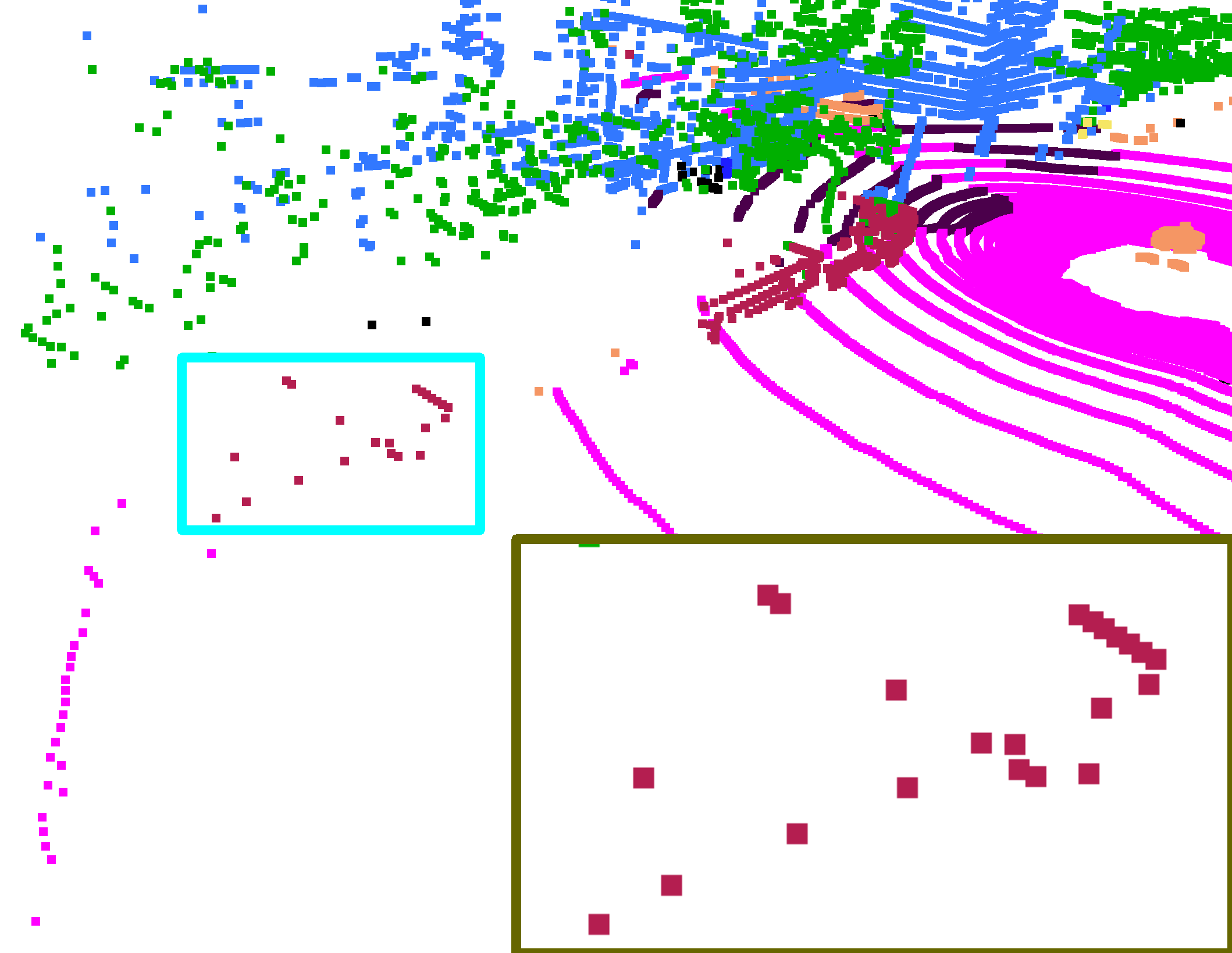}
    \end{minipage}      
     \begin{minipage}  {0.16\linewidth}
        \centering
        \includegraphics [width=1\linewidth,height=0.5\linewidth]
        {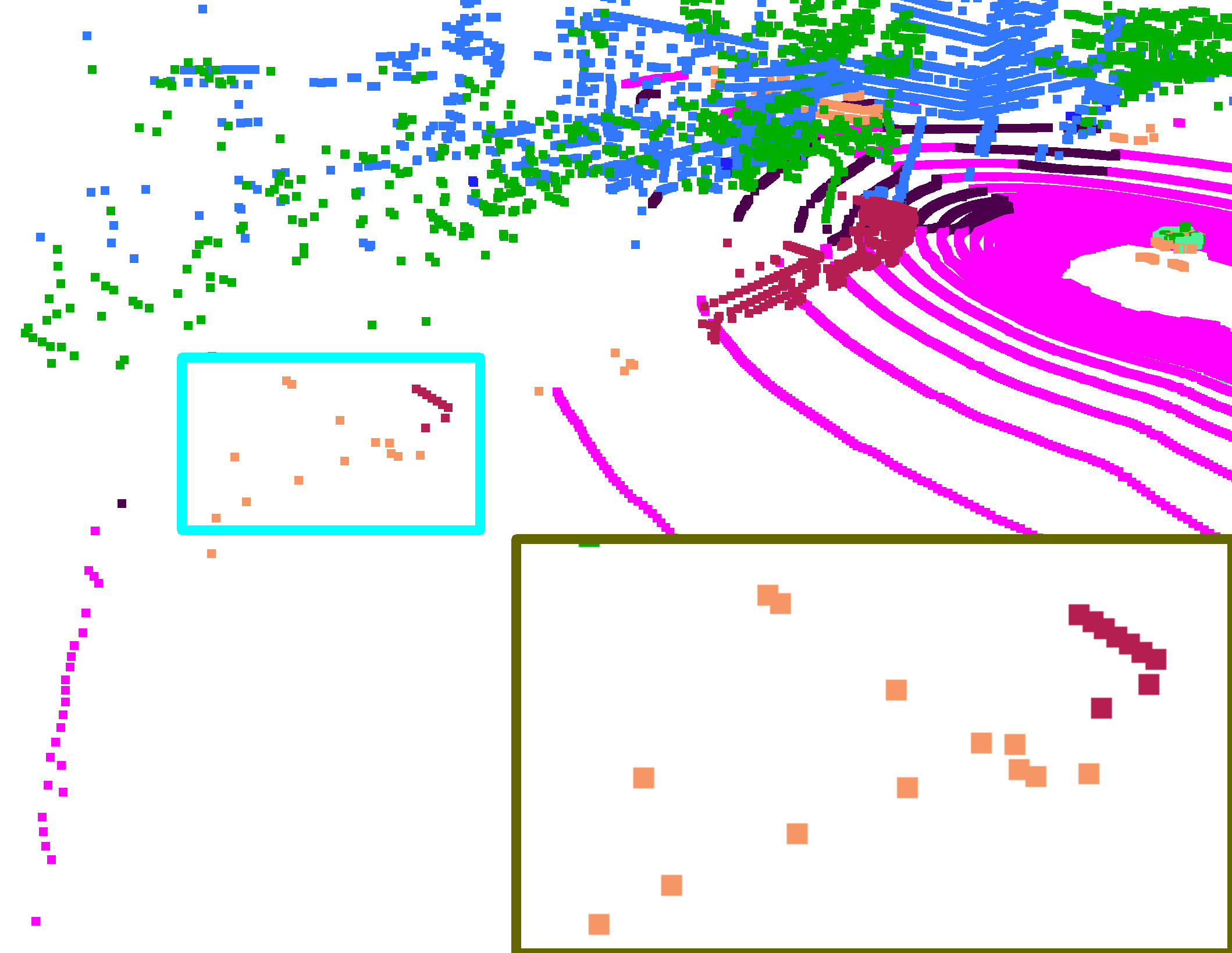}
    \end{minipage} 
    \begin{minipage}  {0.16\linewidth}
        \centering
        \includegraphics [width=1\linewidth,height=0.5\linewidth]
        {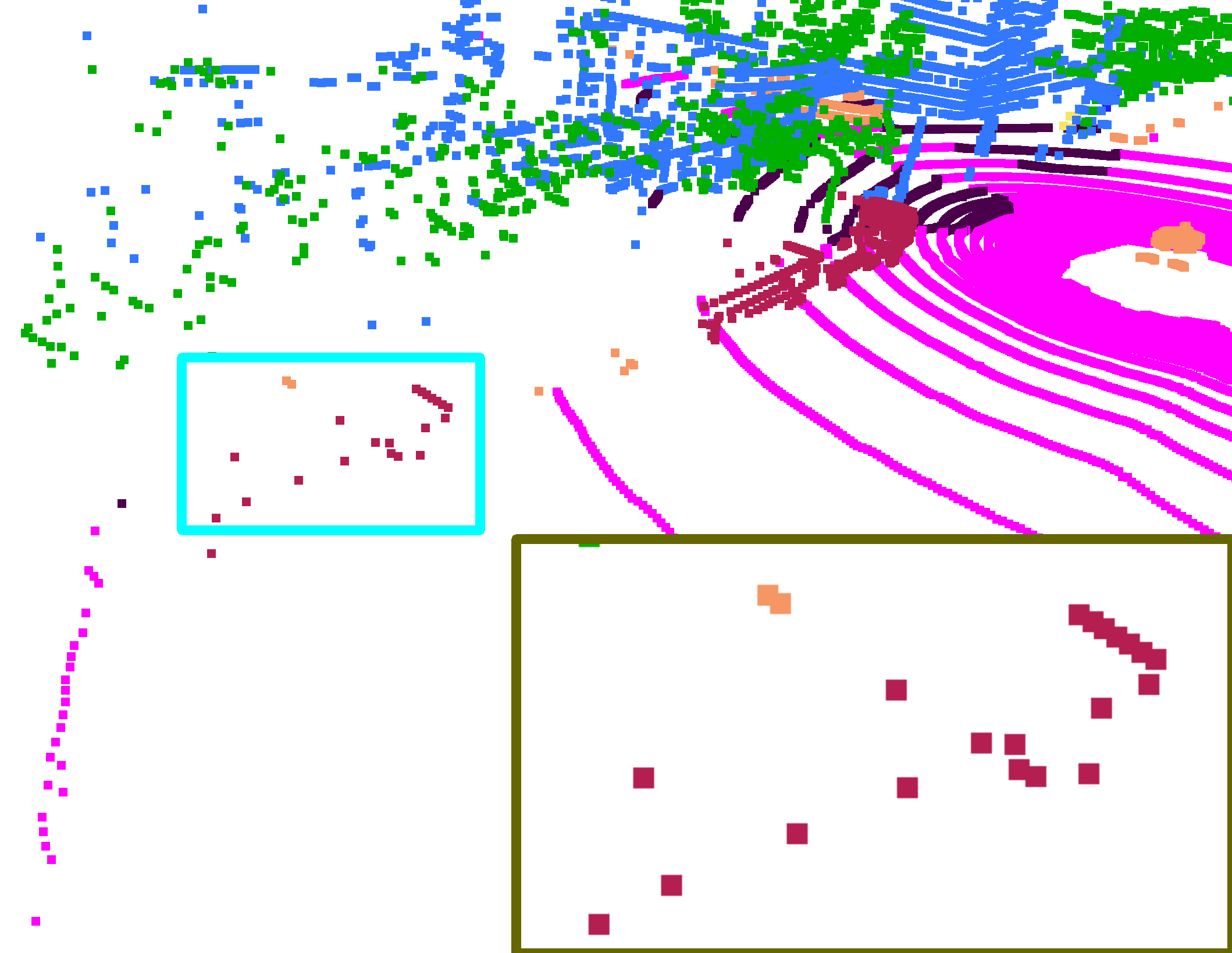}
    \end{minipage}      
     \begin{minipage}  {0.16\linewidth}
        \centering
        \includegraphics [width=1\linewidth,height=0.5\linewidth]
        {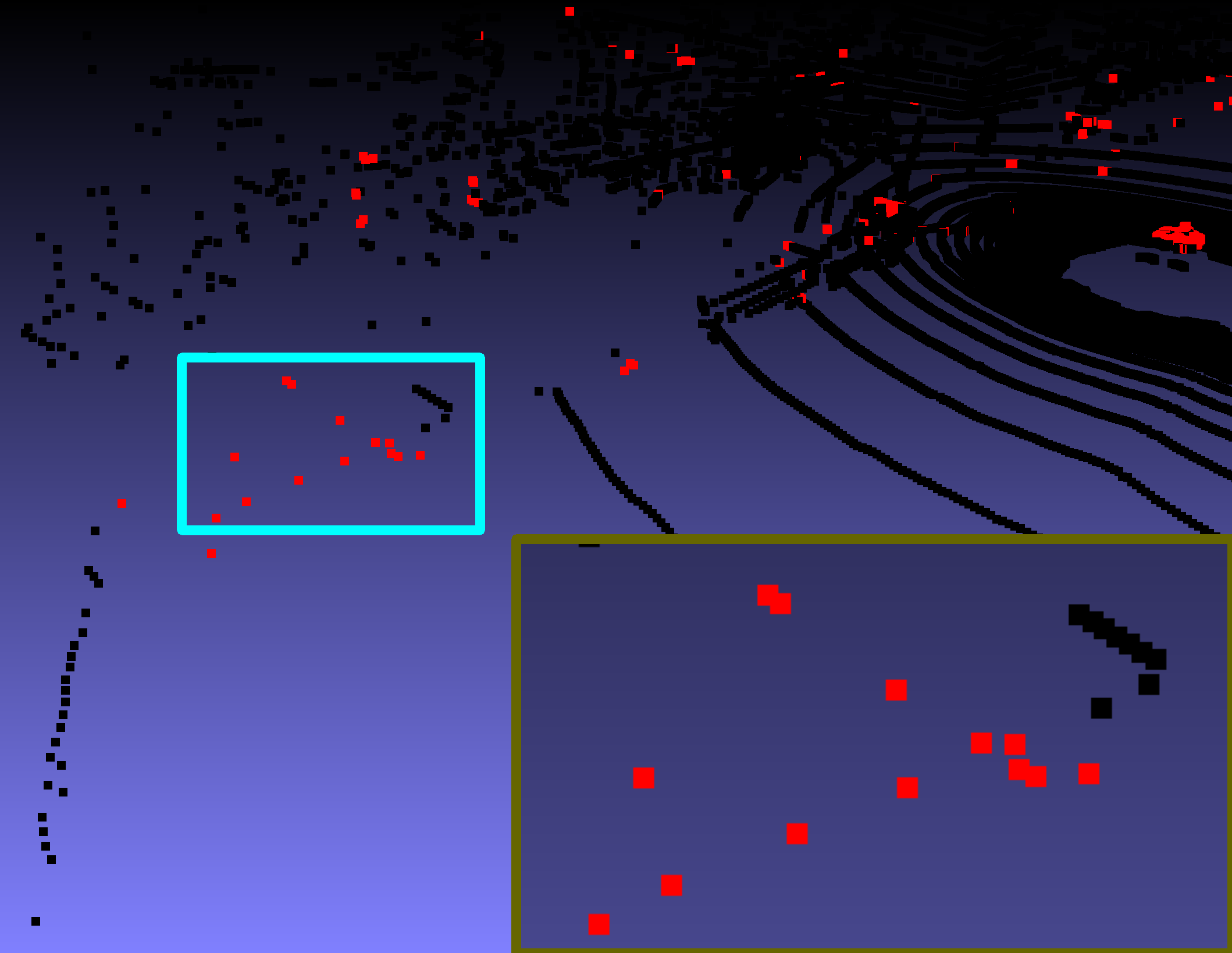}
    \end{minipage} 
     \begin{minipage}  {0.16\linewidth}
        \centering
        \includegraphics [width=1\linewidth,height=0.5\linewidth]
        {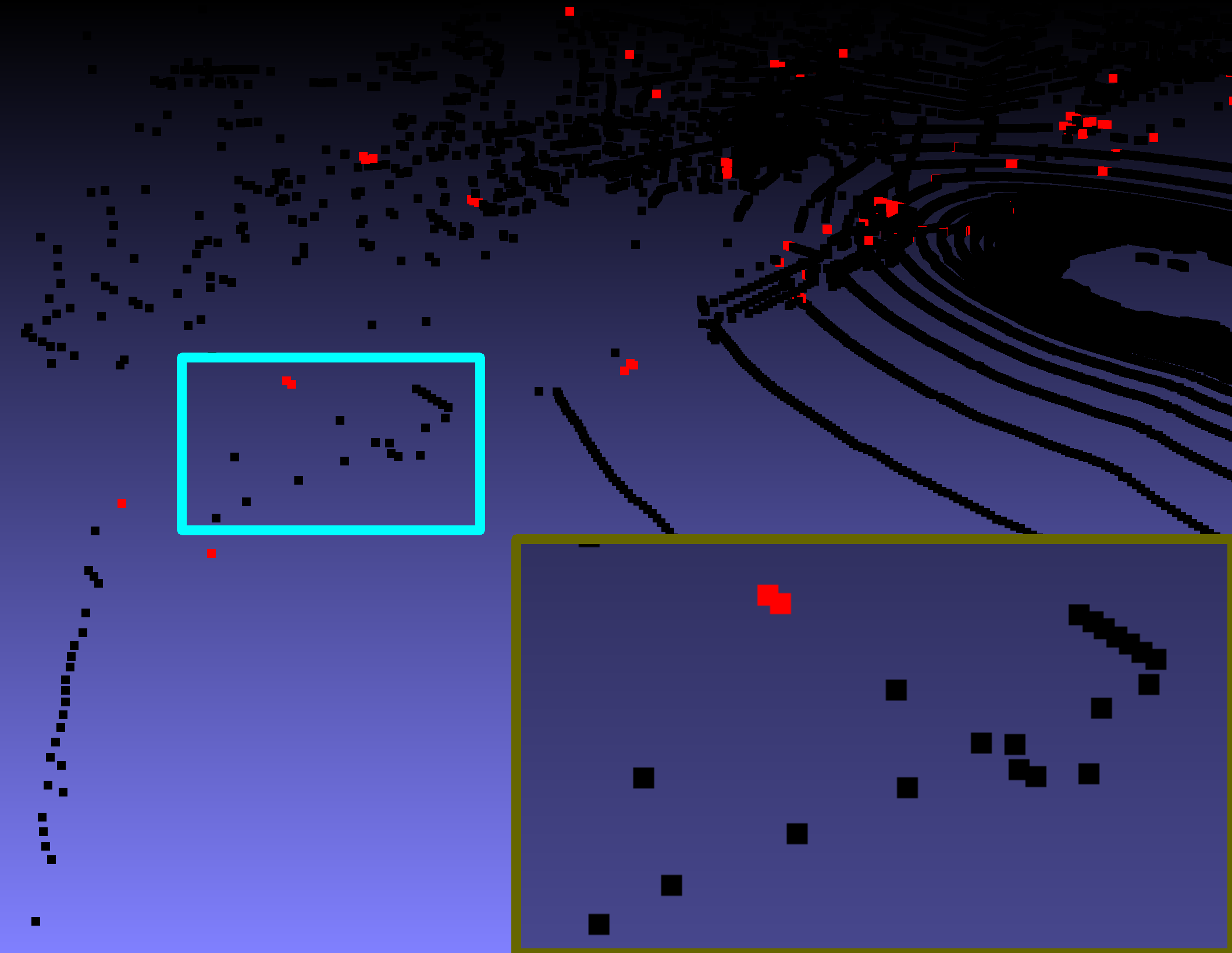}
    \end{minipage} 
	 
	 
	 
    \begin{minipage}  {0.16\linewidth}
        \centering
        \includegraphics [width=1\linewidth,height=0.5\linewidth]
        {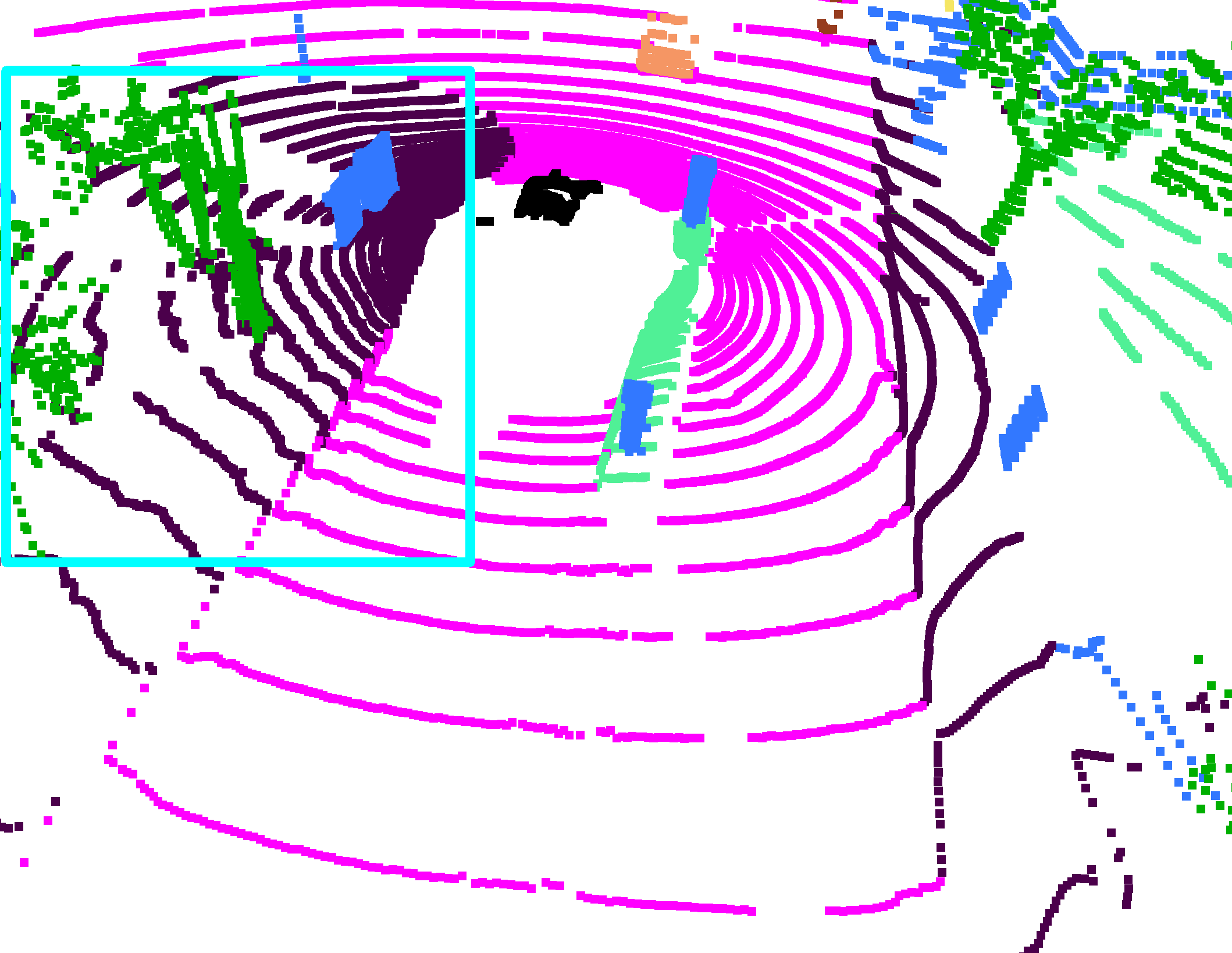}\\\footnotesize{Input}
    \end{minipage}      
    \begin{minipage}  {0.16\linewidth}
        \centering
        \includegraphics [width=1\linewidth,height=0.5\linewidth]
        {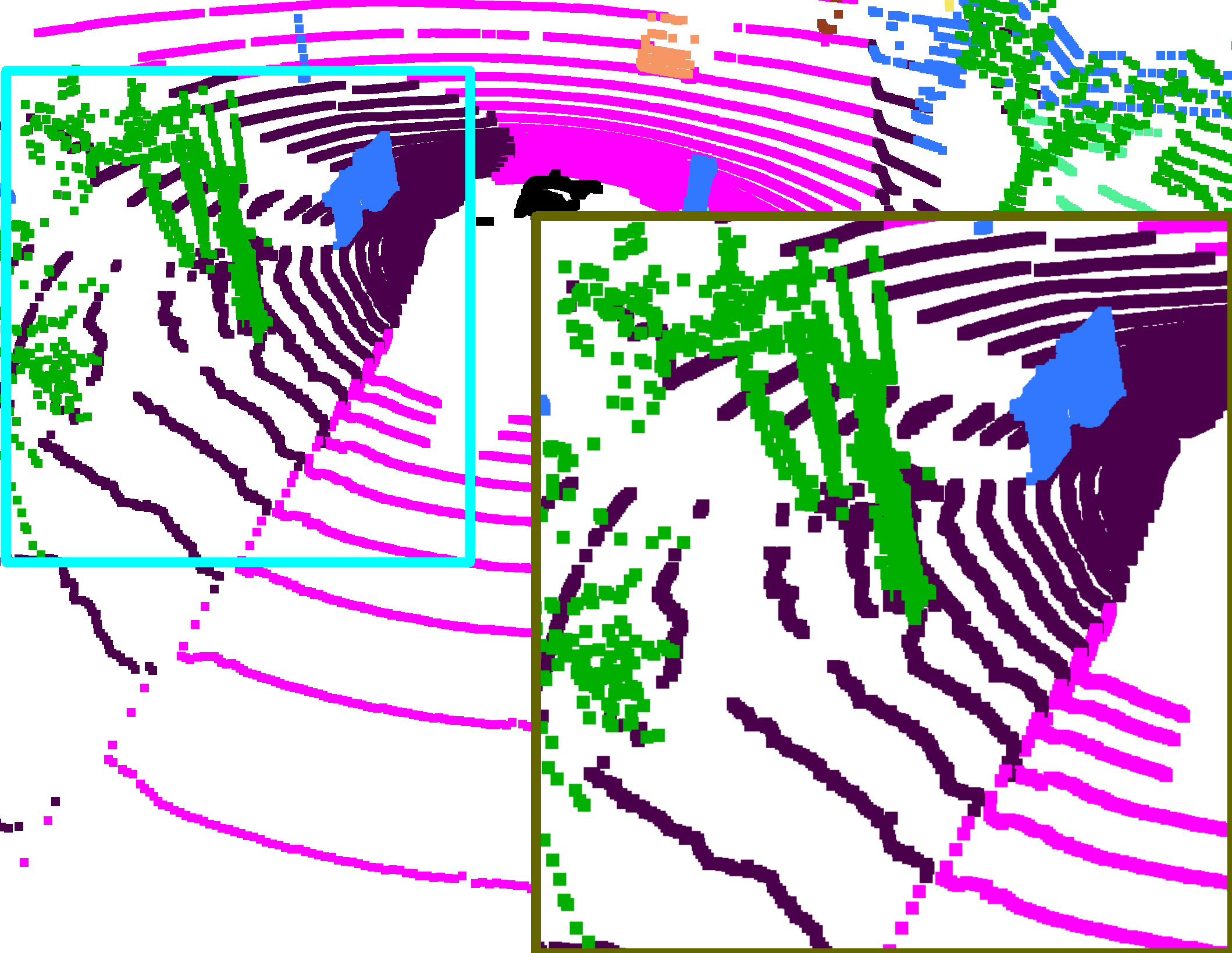}\\\footnotesize{Ground Truth}
    \end{minipage}      
     \begin{minipage}  {0.16\linewidth}
        \centering
        \includegraphics [width=1\linewidth,height=0.5\linewidth]
        {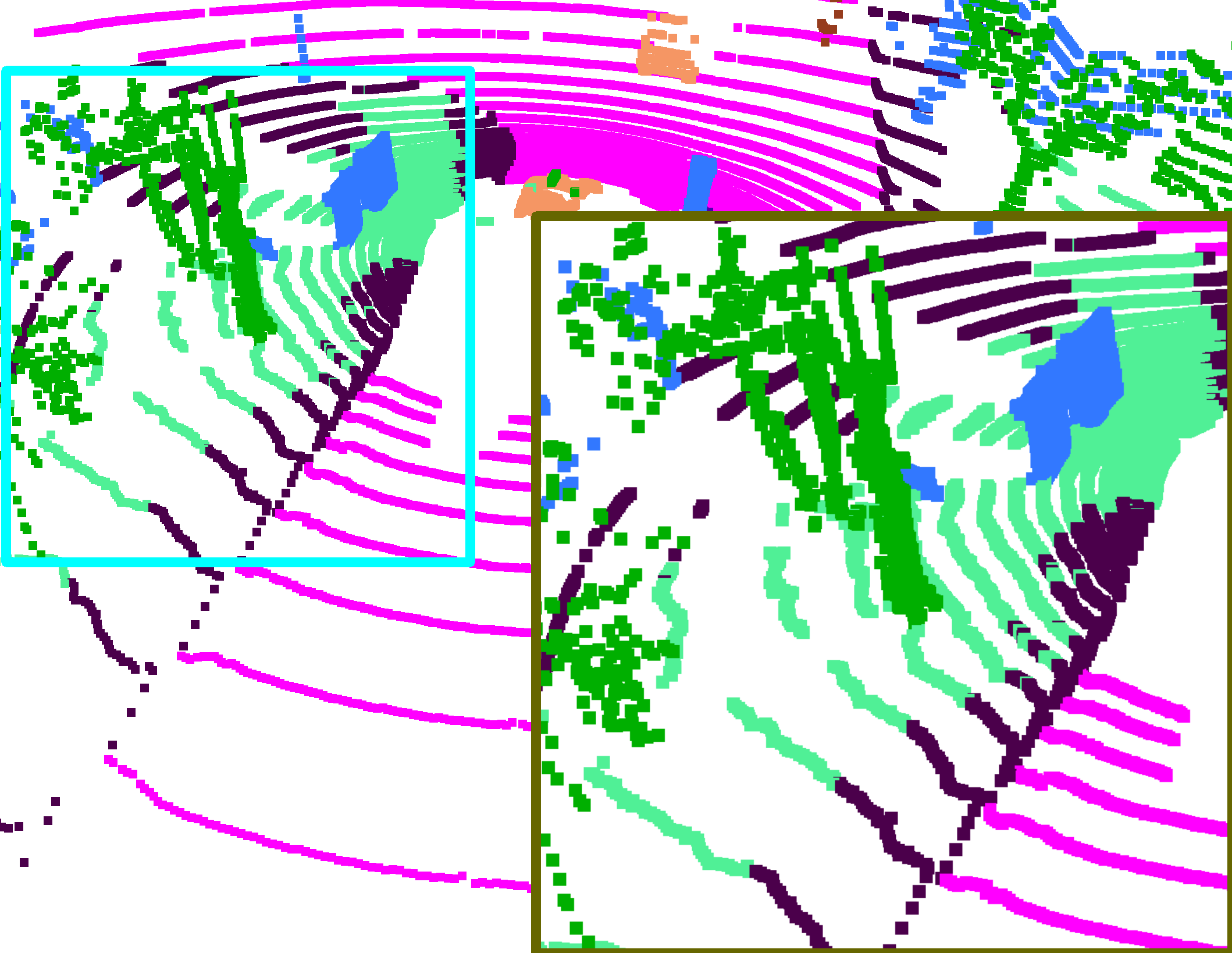}\\\footnotesize{SparseConv}
    \end{minipage} 
    \begin{minipage}  {0.16\linewidth}
        \centering
        \includegraphics [width=1\linewidth,height=0.5\linewidth]
        {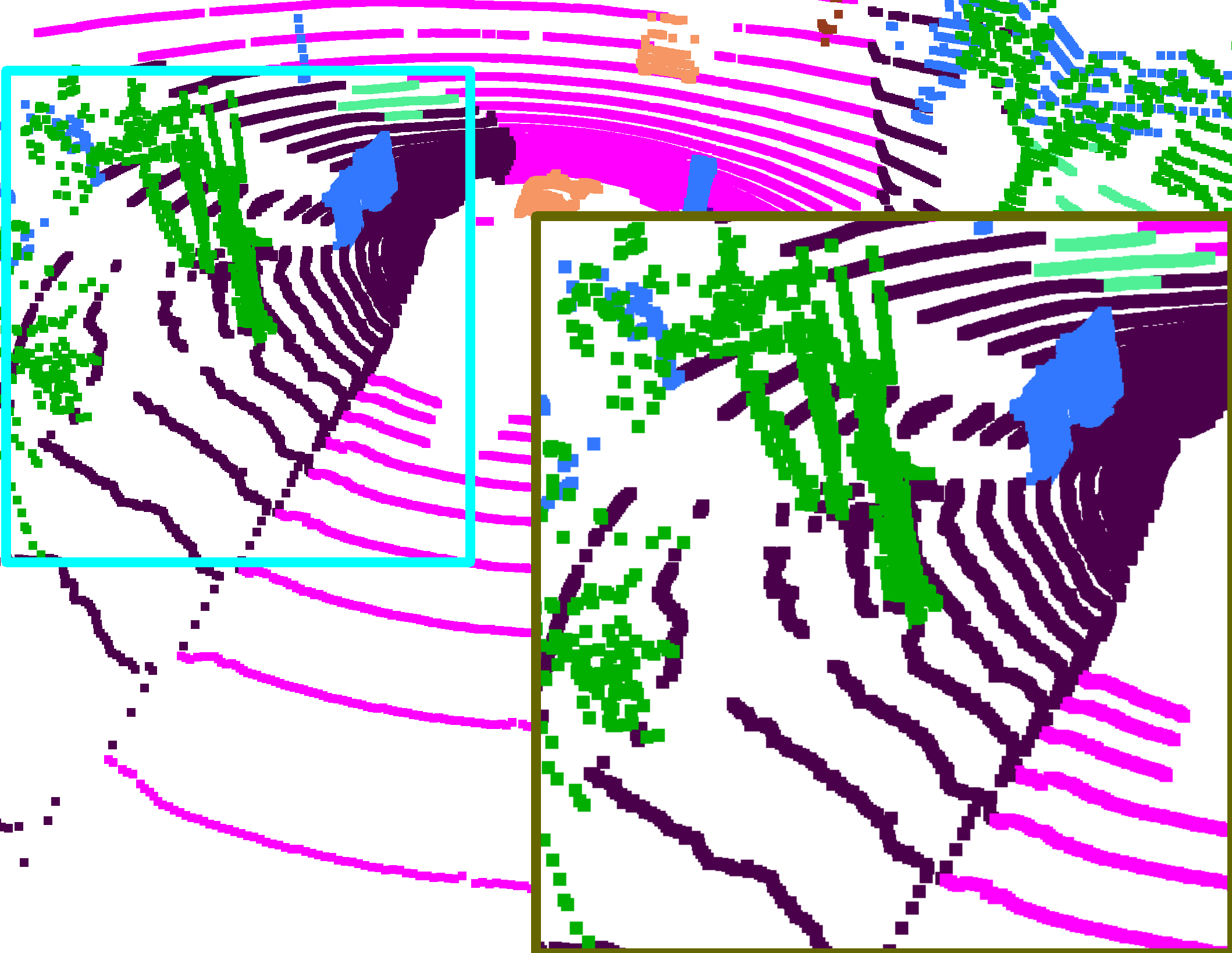}\\\footnotesize{Ours}
    \end{minipage}      
     \begin{minipage}  {0.16\linewidth}
        \centering
        \includegraphics [width=1\linewidth,height=0.5\linewidth]
        {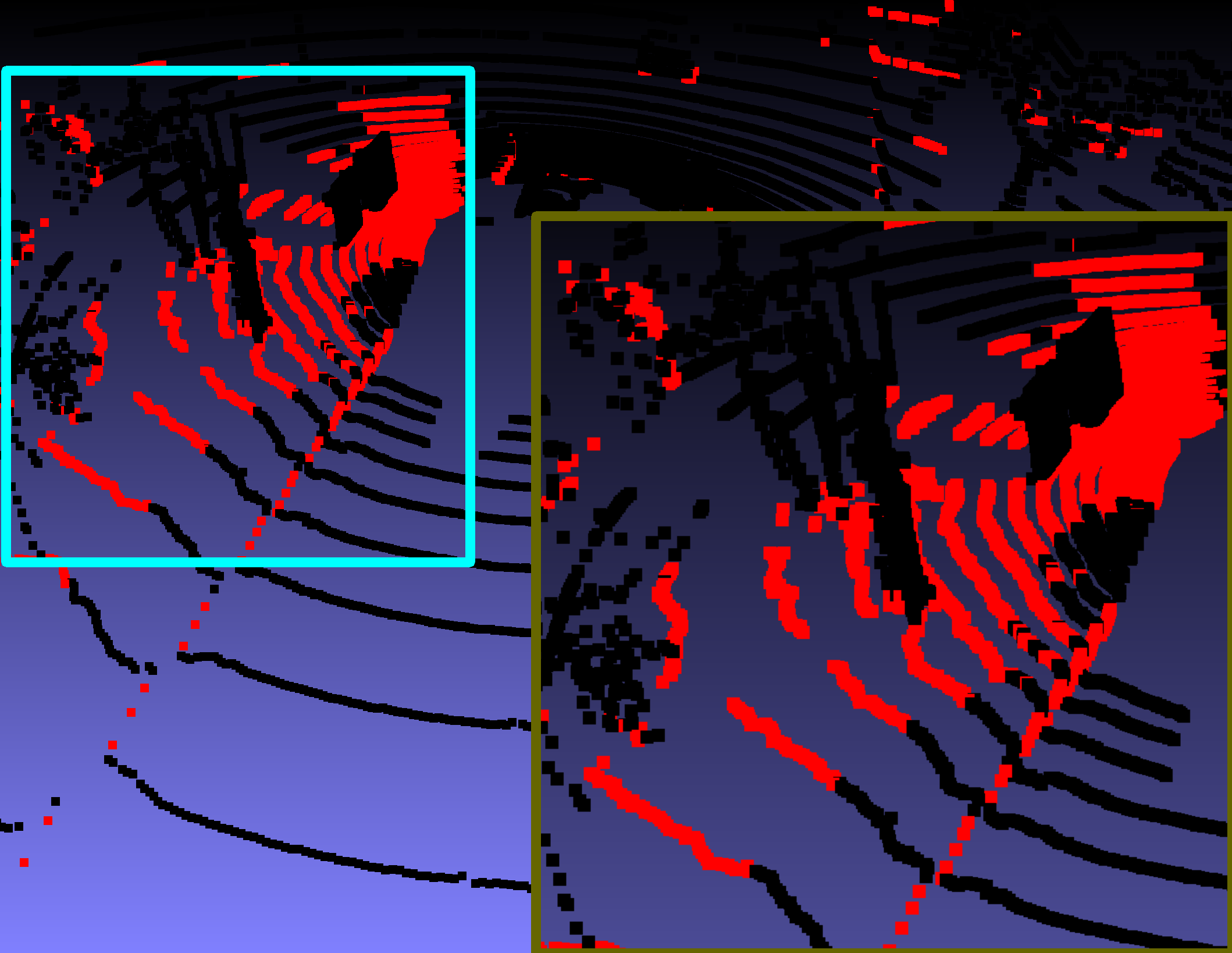}\\\footnotesize{SparseConv}
    \end{minipage} 
     \begin{minipage}  {0.16\linewidth}
        \centering
        \includegraphics [width=1\linewidth,height=0.5\linewidth]
        {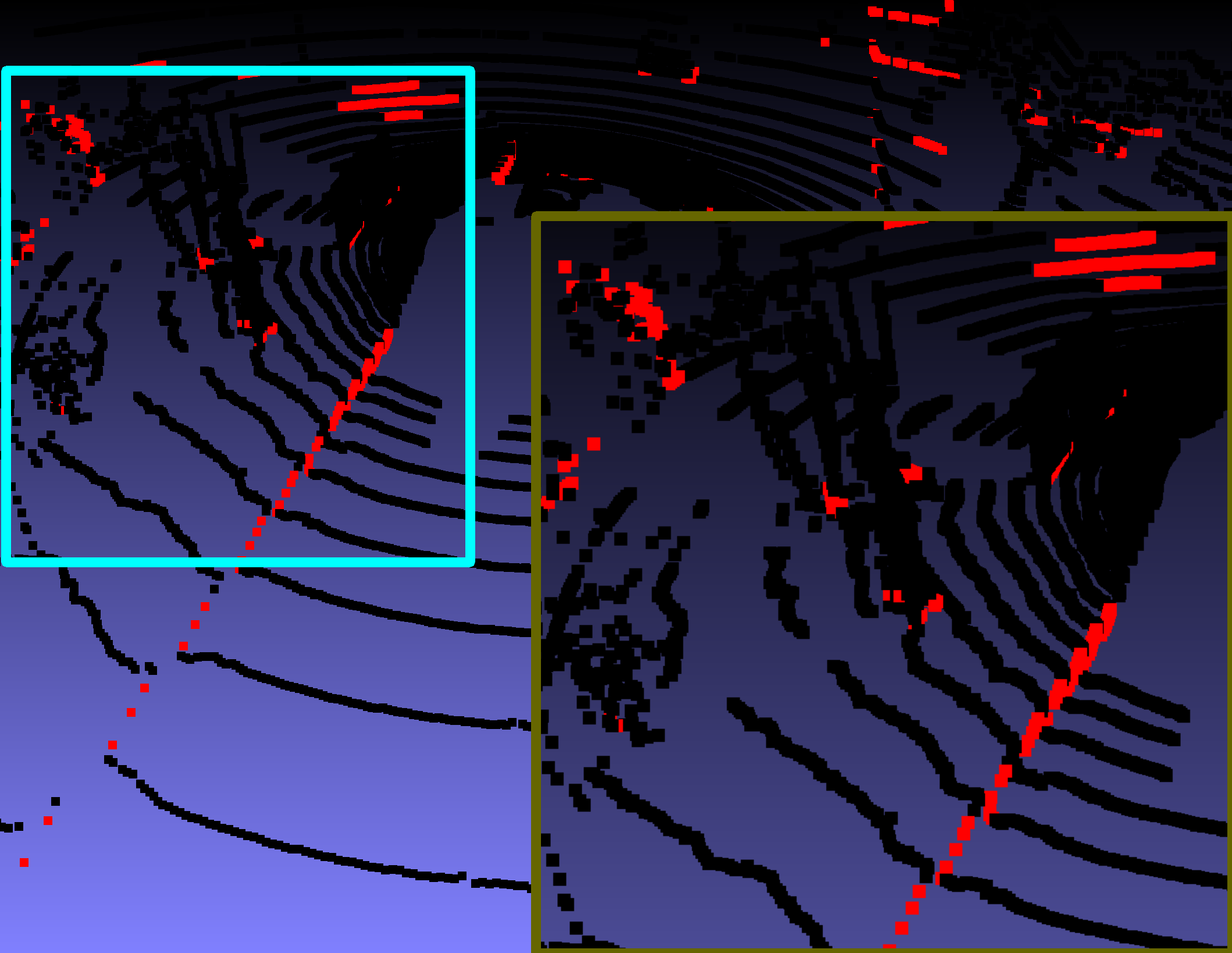}\\\footnotesize{Ours}
    \end{minipage} 
	 
	 \vspace{0.2cm}
	 
    \begin{minipage}  {0.04\linewidth}
        \centering
        \includegraphics [width=0.5\linewidth,height=0.5\linewidth]
        {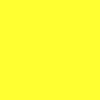}
    \end{minipage}\footnotesize barrier
    \begin{minipage}  {0.04\linewidth}
        \centering
        \includegraphics [width=0.5\linewidth,height=0.5\linewidth]
        {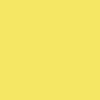}
    \end{minipage}\footnotesize bicycle
    \begin{minipage}  {0.04\linewidth}
        \centering
        \includegraphics [width=0.5\linewidth,height=0.5\linewidth]
        {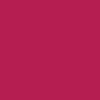}
    \end{minipage}\footnotesize truck
    \begin{minipage}  {0.04\linewidth}
        \centering
        \includegraphics [width=0.5\linewidth,height=0.5\linewidth]
        {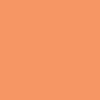}
    \end{minipage}\footnotesize car
    \begin{minipage}  {0.04\linewidth}
        \centering
        \includegraphics [width=0.5\linewidth,height=0.5\linewidth]
        {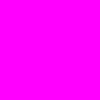}
    \end{minipage}\footnotesize driveable surface
    \begin{minipage}  {0.04\linewidth}
        \centering
        \includegraphics [width=0.5\linewidth,height=0.5\linewidth]
        {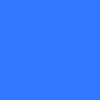}
    \end{minipage}\footnotesize manmade
    \begin{minipage}  {0.04\linewidth}
        \centering
        \includegraphics [width=0.5\linewidth,height=0.5\linewidth]
        {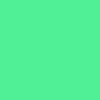}
    \end{minipage}\footnotesize terrain
    \begin{minipage}  {0.04\linewidth}
        \centering
        \includegraphics [width=0.5\linewidth,height=0.5\linewidth]
        {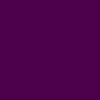}
    \end{minipage}\footnotesize sidewalk
    \begin{minipage}  {0.04\linewidth}
        \centering
        \includegraphics [width=0.5\linewidth,height=0.5\linewidth]
        {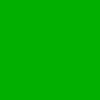}
    \end{minipage}\footnotesize vegetation
	 
    \vspace{-0.1cm}     
    \caption{Visual comparison between vanilla SparseConv and ours (best viewed in color and by zoom-in). The brown box is the zoom-in of the cyan box. The last two columns are the difference maps with the ground truth. More examples are given in the supplementary material.}
    \label{fig:vis_comparison}
    \vspace{-0.2cm}
\end{figure*}

\subsection{Ablation Study}
\label{sec:ablation}
To testify the effectiveness of each component, we conduct an extensive ablation study and list the result in Table~\ref{tab:ablation}. The Experiment~\uppercase\expandafter{\romannumeral1} (Exp.~\uppercase\expandafter{\romannumeral1} for short) is our baseline model of SparseConv. Unless otherwise specified, we train the models on nuScenes \textit{train} set and make evaluations on nuScenes \textit{val} set for the ablation study. To comprehensively 
reveal the effect, we also report the performance at different distances, \ie, close ($\le 20m$), medium ($>20m$ \& $\le 50m$), far ($>50m$) distances.

\paragraph{Window Shape.}
By comparing Experiments~\uppercase\expandafter{\romannumeral1} and \uppercase\expandafter{\romannumeral2} in Table~\ref{tab:ablation}, we can conclude that the radial window shape is beneficial. Further, the improvement stems mainly from better handling the \textit{medium} and \textit{far} points, where we yield $5.67\%$ and $13.39\%$ mIoU performance gain, respectively. This result exactly verifies the benefit of aggregating long-range information with the radial window shape.

Moreover, we also compare the radial window shape with the cubic one proposed in \cite{fan2022embracing,mao2021voxel,lai2022stratified}. As shown in Table~\ref{tab:comp_radial_cubic}, the radial window shape considerably outperforms the cubic one.

Besides, we investigate the effect of window size as shown in Table~\ref{tab:exp_window_size}. Setting it too small may make it hard to capture meaningful information, while setting it too large may increase the optimization difficulty.

\vspace{-0.1cm}
\paragraph{Exponential Splitting.}
Compared to Exp.~\uppercase\expandafter{\romannumeral4}, Exp.~\uppercase\expandafter{\romannumeral5} improves with $1.36\%$ more mIoU, which shows the effectiveness. Moreover, the consistent conclusion could be drawn from Experiments~\uppercase\expandafter{\romannumeral2} and \uppercase\expandafter{\romannumeral3}, where we witness $3.88\%$ and $4.43\%$ more mIoU for the \textit{medium} and \textit{far} points, respectively. Also, we notice that with exponential splitting, all the \textit{close}, \textit{medium}, and \textit{far} points are better dealt with.

\vspace{-0.1cm}
\paragraph{Dynamic Feature Selection.}
From the comparison between Experiments \uppercase\expandafter{\romannumeral3} and \uppercase\expandafter{\romannumeral5}, we note that dynamic feature selection brings a $0.8\%$ mIoU performance gain. Interestingly, we further notice that the gain mainly comes from the \textit{close} points, which indicates that the \textit{close} points may not rely too much on global information, since the dense local information is already enough for correct predictions for the dense close points. It also reveals the fact that points at varying locations should be treated differently. Moreover, the comparison between Exp.~\uppercase\expandafter{\romannumeral2} and \uppercase\expandafter{\romannumeral4} leads to consistent conclusion. Although the performance of \textit{medium} and \textit{far} decreases a little, the \textit{overall} mIoU still increases, since their points number is much than that of the \textit{close} points.


\subsection{Visual Comparison}
\label{sec:vis_comp}
As shown in Fig.~\ref{fig:vis_comparison}, we visually compare the baseline model (\ie, SparseConv) and ours. It visually indicates that with our proposed module, more sparse distant objects are recognized, which are highlighted with cyan boxes. More examples are given in the supplementary material.

\section{Conclusion}

We have studied and dealt with varying-sparsity LiDAR point distribution. We proposed SphereFormer to enable the sparse distant points to directly aggregate information from the close ones. We designed radial window self-attention, which enlarges the receptive field for distant points to intervene with close dense ones. Also, we presented exponential splitting to yield more detailed position encoding. Dynamically selecting local or global features is also helpful. Our method demonstrates powerful performance, ranking 1\textsuperscript{st} on both nuScenes and SemanticKITTI semantic segmentation benchmarks and achieving the 3\textsuperscript{rd} on nuScenes object detection benchmark. It shows a new way to further enhance 3D visual understanding. Our limitations are discussed in the supplementary material.

{\small
\bibliographystyle{ieee_fullname}
\bibliography{egbib}

\begin{thebibliography}{10}\itemsep=-1pt

\bibitem{alonso20203d}
Inigo Alonso, Luis Riazuelo, Luis Montesano, and Ana~C Murillo.
\newblock 3d-mininet: Learning a 2d representation from point clouds for fast
  and efficient 3d lidar semantic segmentation.
\newblock {\em IEEE Robotics and Automation Letters}, 2020.

\bibitem{bai2022transfusion}
Xuyang Bai, Zeyu Hu, Xinge Zhu, Qingqiu Huang, Yilun Chen, Hongbo Fu, and
  Chiew-Lan Tai.
\newblock Transfusion: Robust lidar-camera fusion for 3d object detection with
  transformers.
\newblock In {\em CVPR}, 2022.

\bibitem{behley2019semantickitti}
Jens Behley, Martin Garbade, Andres Milioto, Jan Quenzel, Sven Behnke, Cyrill
  Stachniss, and Jurgen Gall.
\newblock Semantickitti: A dataset for semantic scene understanding of lidar
  sequences.
\newblock In {\em ICCV}, 2019.

\bibitem{caesar2020nuscenes}
Holger Caesar, Varun Bankiti, Alex~H Lang, Sourabh Vora, Venice~Erin Liong,
  Qiang Xu, Anush Krishnan, Yu Pan, Giancarlo Baldan, and Oscar Beijbom.
\newblock nuscenes: A multimodal dataset for autonomous driving.
\newblock In {\em CVPR}, 2020.

\bibitem{detr}
Nicolas Carion, Francisco Massa, Gabriel Synnaeve, Nicolas Usunier, Alexander
  Kirillov, and Sergey Zagoruyko.
\newblock End-to-end object detection with transformers.
\newblock In {\em ECCV}, 2020.

\bibitem{chen2018encoder}
Liang-Chieh Chen, Yukun Zhu, George Papandreou, Florian Schroff, and Hartwig
  Adam.
\newblock Encoder-decoder with atrous separable convolution for semantic image
  segmentation.
\newblock In {\em ECCV}, 2018.

\bibitem{chen2020every}
Qi Chen, Lin Sun, Ernest Cheung, and Alan~L Yuille.
\newblock Every view counts: Cross-view consistency in 3d object detection with
  hybrid-cylindrical-spherical voxelization.
\newblock {\em NeurIPS}, 2020.

\bibitem{chen2020object}
Qi Chen, Lin Sun, Zhixin Wang, Kui Jia, and Alan Yuille.
\newblock Object as hotspots: An anchor-free 3d object detection approach via
  firing of hotspots.
\newblock In {\em ECCV}, 2020.

\bibitem{chen2022focal}
Yukang Chen, Yanwei Li, Xiangyu Zhang, Jian Sun, and Jiaya Jia.
\newblock Focal sparse convolutional networks for 3d object detection.
\newblock In {\em CVPR}, 2022.

\bibitem{chen2022scaling}
Yukang Chen, Jianhui Liu, Xiaojuan Qi, Xiangyu Zhang, Jian Sun, and Jiaya Jia.
\newblock Scaling up kernels in 3d cnns.
\newblock {\em arXiv:2206.10555}, 2022.

\bibitem{chen2023voxenext}
Yukang Chen, Jianhui Liu, Xiangyu Zhang, Xiaojuan Qi, and Jiaya Jia.
\newblock Fully sparse voxelnet for 3d object detection and tracking.
\newblock In {\em CVPR}, 2023.

\bibitem{cheng20212}
Ran Cheng, Ryan Razani, Ehsan Taghavi, Enxu Li, and Bingbing Liu.
\newblock 2-s3net: Attentive feature fusion with adaptive feature selection for
  sparse semantic segmentation network.
\newblock In {\em CVPR}, 2021.

\bibitem{choy20194d}
Christopher Choy, JunYoung Gwak, and Silvio Savarese.
\newblock 4d spatio-temporal convnets: Minkowski convolutional neural networks.
\newblock In {\em CVPR}, 2019.

\bibitem{icm-3d}
Ruihang Chu, Yukang Chen, Tao Kong, Lu Qi, and Lei Li.
\newblock Icm-3d: Instantiated category modeling for 3d instance segmentation.
\newblock {\em IEEE RAL}, 7(1):57--64, 2021.

\bibitem{chu2022twist}
Ruihang Chu, Xiaoqing Ye, Zhengzhe Liu, Xiao Tan, Xiaojuan Qi, Chi-Wing Fu, and
  Jiaya Jia.
\newblock Twist: Two-way inter-label self-training for semi-supervised 3d
  instance segmentation.
\newblock In {\em CVPR}, 2022.

\bibitem{chu2021Twins}
Xiangxiang Chu, Zhi Tian, Yuqing Wang, Bo Zhang, Haibing Ren, Xiaolin Wei,
  Huaxia Xia, and Chunhua Shen.
\newblock Twins: Revisiting the design of spatial attention in vision
  transformers.
\newblock {\em arXiv:2104.13840}, 2021.

\bibitem{cohen2018spherical}
Taco~S Cohen, Mario Geiger, Jonas K{\"o}hler, and Max Welling.
\newblock Spherical cnns.
\newblock {\em arXiv preprint arXiv:1801.10130}, 2018.

\bibitem{cortinhal2020salsanext}
Tiago Cortinhal, George Tzelepis, and Eren Erdal~Aksoy.
\newblock Salsanext: Fast, uncertainty-aware semantic segmentation of lidar
  point clouds.
\newblock In {\em International Symposium on Visual Computing}, 2020.

\bibitem{voxel-rcnn}
Jiajun Deng, Shaoshuai Shi, Peiwei Li, Wengang Zhou, Yanyong Zhang, and
  Houqiang Li.
\newblock Voxel {R-CNN:} towards high performance voxel-based 3d object
  detection.
\newblock In {\em AAAI}, 2021.

\bibitem{dong2021cswin}
Xiaoyi Dong, Jianmin Bao, Dongdong Chen, Weiming Zhang, Nenghai Yu, Lu Yuan,
  Dong Chen, and Baining Guo.
\newblock Cswin transformer: A general vision transformer backbone with
  cross-shaped windows.
\newblock {\em arXiv:2107.00652}, 2021.

\bibitem{dosovitskiy2020vit}
Alexey Dosovitskiy, Lucas Beyer, Alexander Kolesnikov, Dirk Weissenborn,
  Xiaohua Zhai, Thomas Unterthiner, Mostafa Dehghani, Matthias Minderer, Georg
  Heigold, Sylvain Gelly, Jakob Uszkoreit, and Neil Houlsby.
\newblock An image is worth 16x16 words: Transformers for image recognition at
  scale.
\newblock {\em ICLR}, 2021.

\bibitem{fan2022embracing}
Lue Fan, Ziqi Pang, Tianyuan Zhang, Yu-Xiong Wang, Hang Zhao, Feng Wang, Naiyan
  Wang, and Zhaoxiang Zhang.
\newblock {Embracing Single Stride 3D Object Detector with Sparse Transformer}.
\newblock In {\em CVPR}, 2022.

\bibitem{genova2021learning}
Kyle Genova, Xiaoqi Yin, Abhijit Kundu, Caroline Pantofaru, Forrester Cole,
  Avneesh Sud, Brian Brewington, Brian Shucker, and Thomas Funkhouser.
\newblock Learning 3d semantic segmentation with only 2d image supervision.
\newblock In {\em 3DV}, 2021.

\bibitem{3DSemanticSegmentationWithSubmanifoldSparseConvNet}
Benjamin Graham, Martin Engelcke, and Laurens van~der Maaten.
\newblock 3d semantic segmentation with submanifold sparse convolutional
  networks.
\newblock In {\em CVPR}, 2018.

\bibitem{SubmanifoldSparseConvNet}
Benjamin Graham and Laurens van~der Maaten.
\newblock Submanifold sparse convolutional networks.
\newblock {\em arXiv:1706.01307}, 2017.

\bibitem{sassd}
Chenhang He, Hui Zeng, Jianqiang Huang, Xian{-}Sheng Hua, and Lei Zhang.
\newblock Structure aware single-stage 3d object detection from point cloud.
\newblock In {\em CVPR}, pages 11870--11879, 2020.

\bibitem{hou2022point}
Yuenan Hou, Xinge Zhu, Yuexin Ma, Chen~Change Loy, and Yikang Li.
\newblock Point-to-voxel knowledge distillation for lidar semantic
  segmentation.
\newblock In {\em CVPR}, 2022.

\bibitem{hu2020randla}
Qingyong Hu, Bo Yang, Linhai Xie, Stefano Rosa, Yulan Guo, Zhihua Wang, Niki
  Trigoni, and Andrew Markham.
\newblock Randla-net: Efficient semantic segmentation of large-scale point
  clouds.
\newblock In {\em CVPR}, 2020.

\bibitem{jiang2021guided}
Li Jiang, Shaoshuai Shi, Zhuotao Tian, Xin Lai, Shu Liu, Chi-Wing Fu, and Jiaya
  Jia.
\newblock Guided point contrastive learning for semi-supervised point cloud
  semantic segmentation.
\newblock In {\em ICCV}, 2021.

\bibitem{lai2022stratified}
Xin Lai, Jianhui Liu, Li Jiang, Liwei Wang, Hengshuang Zhao, Shu Liu, Xiaojuan
  Qi, and Jiaya Jia.
\newblock Stratified transformer for 3d point cloud segmentation.
\newblock In {\em CVPR}, 2022.

\bibitem{lai2021semi}
Xin Lai, Zhuotao Tian, Li Jiang, Shu Liu, Hengshuang Zhao, Liwei Wang, and
  Jiaya Jia.
\newblock Semi-supervised semantic segmentation with directional context-aware
  consistency.
\newblock In {\em CVPR}, 2021.

\bibitem{lai2022decouplenet}
Xin Lai, Zhuotao Tian, Xiaogang Xu, Yingcong Chen, Shu Liu, Hengshuang Zhao,
  Liwei Wang, and Jiaya Jia.
\newblock Decouplenet: Decoupled network for domain adaptive semantic
  segmentation.
\newblock In {\em ECCV}, 2022.

\bibitem{lang2019pointpillars}
Alex~H Lang, Sourabh Vora, Holger Caesar, Lubing Zhou, Jiong Yang, and Oscar
  Beijbom.
\newblock Pointpillars: Fast encoders for object detection from point clouds.
\newblock In {\em CVPR}, 2019.

\bibitem{li2021simultaneous}
Yiming Li, Tao Kong, Ruihang Chu, Yifeng Li, Peng Wang, and Lei Li.
\newblock Simultaneous semantic and collision learning for 6-dof grasp pose
  estimation.
\newblock In {\em IROS}. IEEE, 2021.

\bibitem{liong2020amvnet}
Venice~Erin Liong, Thi Ngoc~Tho Nguyen, Sergi Widjaja, Dhananjai Sharma, and
  Zhuang~Jie Chong.
\newblock Amvnet: Assertion-based multi-view fusion network for lidar semantic
  segmentation.
\newblock {\em arXiv:2012.04934}, 2020.

\bibitem{spatial-pruned-conv}
Jianhui Liu, Yukang Chen, Xiaoqing Ye, Zhuotao Tian, Xiao Tan, and Xiaojuan Qi.
\newblock Spatial pruned sparse convolution for efficient 3d object detection.
\newblock In {\em NeurIPS}, 2022.

\bibitem{liu2022less}
Minghua Liu, Yin Zhou, Charles~R Qi, Boqing Gong, Hao Su, and Dragomir
  Anguelov.
\newblock Less: Label-efficient semantic segmentation for lidar point clouds.
\newblock In {\em ECCV}, 2022.

\bibitem{liu2021Swin}
Ze Liu, Yutong Lin, Yue Cao, Han Hu, Yixuan Wei, Zheng Zhang, Stephen Lin, and
  Baining Guo.
\newblock Swin transformer: Hierarchical vision transformer using shifted
  windows.
\newblock {\em ICCV}, 2021.

\bibitem{loshchilov2017decoupled}
Ilya Loshchilov and Frank Hutter.
\newblock Decoupled weight decay regularization.
\newblock {\em arXiv:1711.05101}, 2017.

\bibitem{luo2016understanding}
Wenjie Luo, Yujia Li, Raquel Urtasun, and Richard Zemel.
\newblock Understanding the effective receptive field in deep convolutional
  neural networks.
\newblock In {\em NeurIPS}, 2016.

\bibitem{pyramid-rcnn}
Jiageng Mao, Minzhe Niu, Haoyue Bai, Xiaodan Liang, Hang Xu, and Chunjing Xu.
\newblock Pyramid {R-CNN:} towards better performance and adaptability for 3d
  object detection.
\newblock In {\em ICCV}, 2021.

\bibitem{mao2021voxel}
Jiageng Mao, Yujing Xue, Minzhe Niu, Haoyue Bai, Jiashi Feng, Xiaodan Liang,
  Hang Xu, and Chunjing Xu.
\newblock Voxel transformer for 3d object detection.
\newblock In {\em ICCV}, 2021.

\bibitem{milioto2019rangenet++}
Andres Milioto, Ignacio Vizzo, Jens Behley, and Cyrill Stachniss.
\newblock Rangenet++: Fast and accurate lidar semantic segmentation.
\newblock In {\em 2019 IEEE/RSJ international conference on intelligent robots
  and systems (IROS)}, 2019.

\bibitem{qi2017pointnet}
Charles~R Qi, Hao Su, Kaichun Mo, and Leonidas~J Guibas.
\newblock Pointnet: Deep learning on point sets for 3d classification and
  segmentation.
\newblock In {\em CVPR}, 2017.

\bibitem{qi2017pointnet++}
Charles~Ruizhongtai Qi, Li Yi, Hao Su, and Leonidas~J Guibas.
\newblock Pointnet++: Deep hierarchical feature learning on point sets in a
  metric space.
\newblock {\em NeurIPS}, 2017.

\bibitem{razani2021lite}
Ryan Razani, Ran Cheng, Ehsan Taghavi, and Liu Bingbing.
\newblock Lite-hdseg: Lidar semantic segmentation using lite harmonic dense
  convolutions.
\newblock In {\em ICRA}, 2021.

\bibitem{fasterrcnn}
Shaoqing Ren, Kaiming He, Ross~B. Girshick, and Jian Sun.
\newblock Faster {R-CNN:} towards real-time object detection with region
  proposal networks.
\newblock In {\em NeurIPS}, 2015.

\bibitem{Robert_2022_CVPR}
Damien Robert, Bruno Vallet, and Loic Landrieu.
\newblock Learning multi-view aggregation in the wild for large-scale 3d
  semantic segmentation.
\newblock In {\em CVPR}, 2022.

\bibitem{ronneberger2015u}
Olaf Ronneberger, Philipp Fischer, and Thomas Brox.
\newblock U-net: Convolutional networks for biomedical image segmentation.
\newblock In {\em MICCAI}, 2015.

\bibitem{pvrcnn}
Shaoshuai Shi, Chaoxu Guo, Li Jiang, Zhe Wang, Jianping Shi, Xiaogang Wang, and
  Hongsheng Li.
\newblock {PV-RCNN:} point-voxel feature set abstraction for 3d object
  detection.
\newblock In {\em CVPR}, pages 10526--10535, 2020.

\bibitem{point-rcnn}
Shaoshuai Shi, Xiaogang Wang, and Hongsheng Li.
\newblock Pointrcnn: 3d object proposal generation and detection from point
  cloud.
\newblock In {\em CVPR}, pages 770--779, 2019.

\bibitem{sun2020scalability}
Pei Sun, Henrik Kretzschmar, Xerxes Dotiwalla, Aurelien Chouard, Vijaysai
  Patnaik, Paul Tsui, James Guo, Yin Zhou, Yuning Chai, Benjamin Caine, et~al.
\newblock Scalability in perception for autonomous driving: Waymo open dataset.
\newblock In {\em CVPR}, 2020.

\bibitem{sun2022swformer}
Pei Sun, Mingxing Tan, Weiyue Wang, Chenxi Liu, Fei Xia, Zhaoqi Leng, and
  Dragomir Anguelov.
\newblock Swformer: Sparse window transformer for 3d object detection in point
  clouds.
\newblock {\em ECCV}, 2022.

\bibitem{vip}
Shuyang Sun, Xiaoyu Yue, Song Bai, and Philip Torr.
\newblock Visual parser: Representing part-whole hierarchies with transformers.
\newblock {\em arXiv:2107.05790}, 2021.

\bibitem{tang2020searching}
Haotian Tang, Zhijian Liu, Shengyu Zhao, Yujun Lin, Ji Lin, Hanrui Wang, and
  Song Han.
\newblock Searching efficient 3d architectures with sparse point-voxel
  convolution.
\newblock In {\em ECCV}, 2020.

\bibitem{tatarchenko2018tangent}
Maxim Tatarchenko, Jaesik Park, Vladlen Koltun, and Qian-Yi Zhou.
\newblock Tangent convolutions for dense prediction in 3d.
\newblock In {\em CVPR}, 2018.

\bibitem{openpcdet2020}
OpenPCDet~Development Team.
\newblock Openpcdet: An open-source toolbox for 3d object detection from point
  clouds.
\newblock \url{https://github.com/open-mmlab/OpenPCDet}, 2020.

\bibitem{thomas2019kpconv}
Hugues Thomas, Charles~R Qi, Jean-Emmanuel Deschaud, Beatriz Marcotegui,
  Fran{\c{c}}ois Goulette, and Leonidas~J Guibas.
\newblock Kpconv: Flexible and deformable convolution for point clouds.
\newblock In {\em ICCV}, 2019.

\bibitem{tian2022adaptive}
Zhuotao Tian, Pengguang Chen, Xin Lai, Li Jiang, Shu Liu, Hengshuang Zhao, Bei
  Yu, Ming-Chang Yang, and Jiaya Jia.
\newblock Adaptive perspective distillation for semantic segmentation.
\newblock {\em T-PAMI}, 2022.

\bibitem{tian2023learning}
Zhuotao Tian, Jiequan Cui, Li Jiang, Xiaojuan Qi, Xin Lai, Yixin Chen, Shu Liu,
  and Jiaya Jia.
\newblock Learning context-aware classifier for semantic segmentation.
\newblock {\em AAAI}, 2023.

\bibitem{tian2022generalized}
Zhuotao Tian, Xin Lai, Li Jiang, Shu Liu, Michelle Shu, Hengshuang Zhao, and
  Jiaya Jia.
\newblock Generalized few-shot semantic segmentation.
\newblock In {\em CVPR}, 2022.

\bibitem{pmlr-v139-touvron21a}
Hugo Touvron, Matthieu Cord, Matthijs Douze, Francisco Massa, Alexandre
  Sablayrolles, and Herve Jegou.
\newblock Training data-efficient image transformers \& distillation through
  attention.
\newblock In {\em ICML}, 2021.

\bibitem{touvron2021cait}
Hugo Touvron, Matthieu Cord, Alexandre Sablayrolles, Gabriel Synnaeve, and
  Herv\'e J\'egou.
\newblock Going deeper with image transformers.
\newblock {\em arXiv:2103.17239}, 2021.

\bibitem{vaswani2017attention}
Ashish Vaswani, Noam Shazeer, Niki Parmar, Jakob Uszkoreit, Llion Jones,
  Aidan~N Gomez, {\L}ukasz Kaiser, and Illia Polosukhin.
\newblock Attention is all you need.
\newblock In {\em NeurIPS}, 2017.

\bibitem{wang2021pvtv2}
Wenhai Wang, Enze Xie, Xiang Li, Deng-Ping Fan, Kaitao Song, Ding Liang, Tong
  Lu, Ping Luo, and Ling Shao.
\newblock Pvtv2: Improved baselines with pyramid vision transformer.
\newblock {\em arXiv:2106.13797}, 2021.

\bibitem{wang2021pyramid}
Wenhai Wang, Enze Xie, Xiang Li, Deng-Ping Fan, Kaitao Song, Ding Liang, Tong
  Lu, Ping Luo, and Ling Shao.
\newblock Pyramid vision transformer: A versatile backbone for dense prediction
  without convolutions.
\newblock In {\em ICCV}, 2021.

\bibitem{wu2019squeezesegv2}
Bichen Wu, Xuanyu Zhou, Sicheng Zhao, Xiangyu Yue, and Kurt Keutzer.
\newblock Squeezesegv2: Improved model structure and unsupervised domain
  adaptation for road-object segmentation from a lidar point cloud.
\newblock In {\em ICRA}, 2019.

\bibitem{xu2020squeezesegv3}
Chenfeng Xu, Bichen Wu, Zining Wang, Wei Zhan, Peter Vajda, Kurt Keutzer, and
  Masayoshi Tomizuka.
\newblock Squeezesegv3: Spatially-adaptive convolution for efficient
  point-cloud segmentation.
\newblock In {\em ECCV}, 2020.

\bibitem{xu2021rpvnet}
Jianyun Xu, Ruixiang Zhang, Jian Dou, Yushi Zhu, Jie Sun, and Shiliang Pu.
\newblock Rpvnet: A deep and efficient range-point-voxel fusion network for
  lidar point cloud segmentation.
\newblock In {\em ICCV}, 2021.

\bibitem{yan2021sparse}
Xu Yan, Jiantao Gao, Jie Li, Ruimao Zhang, Zhen Li, Rui Huang, and Shuguang
  Cui.
\newblock Sparse single sweep lidar point cloud segmentation via learning
  contextual shape priors from scene completion.
\newblock In {\em AAAI}, 2021.

\bibitem{yan20222dpass}
Xu Yan, Jiantao Gao, Chaoda Zheng, Chao Zheng, Ruimao Zhang, Shuguang Cui, and
  Zhen Li.
\newblock 2dpass: 2d priors assisted semantic segmentation on lidar point
  clouds.
\newblock In {\em ECCV}, 2022.

\bibitem{yan2020pointasnl}
Xu Yan, Chaoda Zheng, Zhen Li, Sheng Wang, and Shuguang Cui.
\newblock Pointasnl: Robust point clouds processing using nonlocal neural
  networks with adaptive sampling.
\newblock In {\em CVPR}, 2020.

\bibitem{second}
Yan Yan, Yuxing Mao, and Bo Li.
\newblock {SECOND:} sparsely embedded convolutional detection.
\newblock {\em Sensors}, 18(10):3337, 2018.

\bibitem{yang2021focal}
Jianwei Yang, Chunyuan Li, Pengchuan Zhang, Xiyang Dai, Bin Xiao, Lu Yuan, and
  Jianfeng Gao.
\newblock Focal self-attention for local-global interactions in vision
  transformers.
\newblock In {\em NeurIPS}, 2021.

\bibitem{3dssd}
Zetong Yang, Yanan Sun, Shu Liu, and Jiaya Jia.
\newblock 3dssd: Point-based 3d single stage object detector.
\newblock In {\em CVPR}, pages 11037--11045, 2020.

\bibitem{yang20203dssd}
Zetong Yang, Yanan Sun, Shu Liu, and Jiaya Jia.
\newblock 3dssd: Point-based 3d single stage object detector.
\newblock In {\em CVPR}, 2020.

\bibitem{ye2022lidarmultinet}
Dongqiangzi Ye, Zixiang Zhou, Weijia Chen, Yufei Xie, Yu Wang, Panqu Wang, and
  Hassan Foroosh.
\newblock Lidarmultinet: Towards a unified multi-task network for lidar
  perception.
\newblock {\em arXiv:2209.09385}, 2022.

\bibitem{yin2021center}
Tianwei Yin, Xingyi Zhou, and Philipp Krahenbuhl.
\newblock Center-based 3d object detection and tracking.
\newblock In {\em CVPR}, 2021.

\bibitem{zhang2020polarnet}
Yang Zhang, Zixiang Zhou, Philip David, Xiangyu Yue, Zerong Xi, Boqing Gong,
  and Hassan Foroosh.
\newblock Polarnet: An improved grid representation for online lidar point
  clouds semantic segmentation.
\newblock In {\em CVPR}, 2020.

\bibitem{zhao2020san}
Hengshuang Zhao, Jiaya Jia, and Vladlen Koltun.
\newblock Exploring self-attention for image recognition.
\newblock In {\em CVPR}, 2020.

\bibitem{zhao2021point}
Hengshuang Zhao, Li Jiang, Jiaya Jia, Philip~HS Torr, and Vladlen Koltun.
\newblock Point transformer.
\newblock In {\em ICCV}, 2021.

\bibitem{zhao2017pyramid}
Hengshuang Zhao, Jianping Shi, Xiaojuan Qi, Xiaogang Wang, and Jiaya Jia.
\newblock Pyramid scene parsing network.
\newblock In {\em CVPR}, 2017.

\bibitem{cia-ssd}
Wu Zheng, Weiliang Tang, Sijin Chen, Li Jiang, and Chi{-}Wing Fu.
\newblock {CIA-SSD:} confident iou-aware single-stage object detector from
  point cloud.
\newblock In {\em AAAI}, pages 3555--3562, 2021.

\bibitem{sessd}
Wu Zheng, Weiliang Tang, Li Jiang, and Chi{-}Wing Fu.
\newblock {SE-SSD:} self-ensembling single-stage object detector from point
  cloud.
\newblock In {\em CVPR}, pages 14494--14503, 2021.

\bibitem{voxelnet}
Yin Zhou and Oncel Tuzel.
\newblock Voxelnet: End-to-end learning for point cloud based 3d object
  detection.
\newblock In {\em CVPR}, pages 4490--4499, 2018.

\bibitem{zhu2019class}
Benjin Zhu, Zhengkai Jiang, Xiangxin Zhou, Zeming Li, and Gang Yu.
\newblock Class-balanced grouping and sampling for point cloud 3d object
  detection.
\newblock {\em arXiv:1908.09492}, 2019.

\bibitem{zhu2020deformable}
Xizhou Zhu, Weijie Su, Lewei Lu, Bin Li, Xiaogang Wang, and Jifeng Dai.
\newblock Deformable detr: Deformable transformers for end-to-end object
  detection.
\newblock In {\em ICLR}, 2020.

\bibitem{zhu2021cylindrical}
Xinge Zhu, Hui Zhou, Tai Wang, Fangzhou Hong, Yuexin Ma, Wei Li, Hongsheng Li,
  and Dahua Lin.
\newblock Cylindrical and asymmetrical 3d convolution networks for lidar
  segmentation.
\newblock In {\em CVPR}, 2021.

\bibitem{zhuang2021perception}
Zhuangwei Zhuang, Rong Li, Kui Jia, Qicheng Wang, Yuanqing Li, and Mingkui Tan.
\newblock Perception-aware multi-sensor fusion for 3d lidar semantic
  segmentation.
\newblock In {\em ICCV}, 2021.

\end{thebibliography}
}

\end{document}